\documentclass{article}

\PassOptionsToPackage{numbers, compress}{natbib}

\usepackage[preprint]{neurips_data_2024}

\usepackage[utf8]{inputenc} %
\usepackage[T1]{fontenc}    %
\usepackage{hyperref}       %
\usepackage{url}            %
\usepackage{doi}            %
\usepackage{booktabs}       %
\usepackage{amsfonts}       %
\usepackage{nicefrac}       %
\usepackage{microtype}      %
\usepackage{xcolor}         %

\usepackage{amsmath}
\usepackage{amssymb}
\usepackage{graphicx}
\usepackage{algorithm2e}
\usepackage{comment}
\usepackage{dirtytalk}
\usepackage{longtable}%
\usepackage{array}
\usepackage{caption} %
\usepackage{adjustbox} %
\usepackage{appendix} %
\newcolumntype{H}{>{\setbox0=\hbox\bgroup}c<{\egroup}@{}}
\usepackage{multirow} %
\usepackage{makecell}

\newcommand{\eg}{\textit{e}.\textit{g}. }
\newcommand{\errfmt}[2]{#1\,(#2)}

\title{ALPEC: A Comprehensive Evaluation Framework and Dataset for Machine Learning-Based Arousal Detection in Clinical Practice}

\author{%
  Stefan Kraft\thanks{Also: University of Tübingen, GER} \\
  IT-Designers Gruppe\\
  Esslingen am Neckar, GER \\
  \texttt{stefan.kraft@stz-softwaretechnik.de} \\
  \And
  Andreas Theissler\\
  Aalen University of Applied Sciences\\
  Aalen, GER\\
  \texttt{https://orcid.org/0000-0003-0746-0424} \\
  \AND
  Vera Wienhausen-Wilke \\
  Klinikum Esslingen, Klinik für Kardiologie, Pneumologie und Angilologie \\
  Esslingen am Neckar, GER \\
  \texttt{v.wienhausen-wilke@klinikum-esslingen.de} \\
  \And
  Philipp Walter \\
  IT-Designers Gruppe\\
  Esslingen am Neckar, GER \\
  \texttt{philipp.walter@it-designers.de} \\
  \And
  Gjergji Kasneci \\
  Technical University of Munich \\
  Munich, GER \\
  \texttt{gjergji.kasneci@tum.de} \\
}

\begin{document}

\maketitle

\definecolor{questionblue}{RGB}{0, 0, 255}

\newcommand{\question}[1]{\textcolor{questionblue}{#1}}

\begin{abstract}
Detecting arousals in sleep is essential for diagnosing sleep disorders. However, using Machine Learning (ML) in clinical practice is impeded by fundamental issues, primarily due to mismatches between clinical protocols and ML methods. Clinicians typically annotate only the onset of arousals, while ML methods rely on annotations for both the beginning and end. Additionally, there is no standardized evaluation methodology tailored to clinical needs for arousal detection models.
This work addresses these issues by introducing a novel post-processing and evaluation framework emphasizing approximate localization and precise event count (ALPEC) of arousals. We recommend that ML practitioners focus on detecting arousal onsets, aligning with clinical practice. We examine the impact of this shift on current training and evaluation schemes, addressing simplifications and challenges. We utilize a novel comprehensive polysomnographic dataset (CPS) that reflects the aforementioned clinical annotation constraints and includes modalities not present in existing polysomnographic datasets. We release the dataset alongside this paper, demonstrating the benefits of leveraging multimodal data for arousal onset detection.
Our findings significantly contribute to integrating ML-based arousal detection in clinical settings, reducing the gap between technological advancements and clinical needs.
\end{abstract}

\section{Introduction}\label{sec:introduction}

Arousals are short-term biological activation processes during sleep and wakefulness that elevate the organism from a lower to a higher state of mental and physical activity~\citep{raschke1997arousal}. 
Frequent arousals during sleep disrupt deep sleep stages and REM sleep, compromising the restorative function of sleep and causing fragmentation~\citep{wetter2012elsevier}. Arousals are indicative of several sleep disorders, with obstructive sleep apnea (OSA) being the most prevalent breathing-related sleep disorder. OSA, characterized by partial or complete obstructions of the upper airway, results in oxygen desaturation and frequent arousals~\citep{wetter2012elsevier}. This disorder is a significant public health concern, with prevalence estimates around 20\% in men and 17\% in women, 
and is linked to severe health outcomes like hypertension, cardiovascular disease, and increased mortality risk~\citep{franklin2015obstructive,wetter2012elsevier,punjabi2008epidemiology}.

Detecting arousals is a routine task in polysomnographic (PSG) examinations, which involve comprehensive recording and analysis of various physiological parameters such as brain waves, blood oxygen levels, breathing, eye and leg movements during sleep, conducted in sleep laboratories.
However, the diversity of equipment, software, and protocols across laboratories poses significant challenges for developing universally applicable Machine Learning (ML) models for arousal detection~\citep{anido2023decentralized}. Even among laboratories using the same equipment, differences in settings and protocols complicate the development of general-purpose ML-based detectors. The lack of large-scale time series datasets, absence of clear evaluation metrics, and limited consensus on theoretical and practical understanding of time series further impede progress~\citep{garza2023timegpt}. A significant drop in performance of current sleep stage classification models on data from different sleep laboratories highlights the need for arousal detection models trained on data specific to the clinical environments where they will be used~\citep{anido2023decentralized}.

According to the American Association of Sleep Medicine (AASM), arousals are defined as abrupt changes in EEG frequency lasting at least 3 seconds following 10 seconds of sleep~\citep{berry2012rules}. Although research suggests that the duration of arousals may have clinical significance~\citep{schwartz2006potential, shahrbabaki2021sleep}, this finding has not been integrated into clinical practice. Our dataset of PSG recordings, released alongside this paper, reflects this gap, with almost all annotated arousals being three seconds long due to the default setting of the clinical annotation software~\citep{kraft2024cps, goldberger2000physiobank}. This limitation means that only the onset annotations are practically useful. Current ML-based arousal detection research, which depends on full annotations including both the start and end of each event, diverges from this clinical practice, thus limiting the applicability of such models in real-world clinical settings.

Our \textbf{first main contribution} is advocating for a shift in focus from full event detection to detecting arousal onsets to better align with clinical needs. We explore the implications on various training methodologies for binary event detection, addressing both simplifications and emerging challenges. Conversely, aligning clinical practices with ML model requirements without evident patient benefits would unnecessarily burden sleep laboratories, increasing operational challenges such as the scarcity of trained scorers and long patient wait times.
Apart from task misalignment, the fragmented landscape of evaluation methodologies is not tailored for operational utility in real-world healthcare settings. We advocate that arousal detection systems should be developed as clinical decision support systems (CDSS) rather than for autonomous decision-making. This approach adheres to ethical standards in healthcare~\citep{fawzy2023ethics} 
and anticipates evolving regulatory requirements, such as the forthcoming EU AI Act, which mandates human oversight for AI systems in critical areas like healthcare~\citep{madiega2021artificial}.

Our \textbf{second main contribution} is devising the first post-processing and performance evaluation framework for arousal detection guided by clinical goals. This framework is embedded in a relevant taxonomy, overcomes conceptual limitations, adheres to best practices, and aims to standardize performance evaluation in the field. We thoroughly compare our framework to existing evaluation methodologies.

Lastly, as our \textbf{third main contribution}, we are excited to release the Comprehensive Polysomnography (CPS) dataset, a curated, feature-rich collection unique for its extensive channels and novel beat-by-beat blood pressure annotations. This dataset aims to advance ML models in sleep disorder diagnostics. We demonstrate enhanced arousal detection capabilities through multimodal data in our dataset, advancing the integration of previously underutilized data modalities.

\section{Related work}\label{sec:related_work}

\begin{table}[htbp]
\caption{\textbf{Comparison of Related Work}. This comparison showcases the diversity of methodologies utilized in the field. Notations: \textit{2018 Phys} refers to the 2018 PhysioNet Challenge dataset~\citep{ghassemi2018you, goldberger2000physiobank}; \textit{Seg.} denotes segmentation methods; \textit{(AU)PRC} and \textit{(AU)ROC} indicate that either the curve, the area under the curve, or both are reported.}
\label{tab:related_work}
\setlength{\tabcolsep}{3.3pt} %
\begin{center}
\begin{tabular}{llllllllllllllllllllllll}
\toprule
\multirow{2}{*}{Authors} & \multicolumn{2}{c}{Task} & \multicolumn{2}{c}{Seg.} & \multicolumn{5}{c}{Dataset} &  \multicolumn{9}{c}{Evaluation measures} & \multicolumn{4}{c}{Modalities}\\
\cmidrule(lr){2-3}
\cmidrule(lr){4-5}
\cmidrule(lr){6-10}
\cmidrule(lr){11-19}
\cmidrule(lr){20-23}
& \parbox[t]{2mm}{\rotatebox[origin=c]{90}{Arousal}} & \parbox[t]{2mm}{\rotatebox[origin=c]{90}{Sleep Stage}} & \parbox[t]{2mm}{\rotatebox[origin=c]{90}{Windowed}} & \parbox[t]{2mm}{\rotatebox[origin=c]{90}{Pointwise}} & \parbox[t]{2mm}{\rotatebox[origin=c]{90}{2018 Phys~\citep{ghassemi2018you}}} & \parbox[t]{2mm}{\rotatebox[origin=c]{90}{SHHS~\cite{quan1997sleep}}} & \parbox[t]{2mm}{\rotatebox[origin=c]{90}{MESA~\citep{chen2015racial}}} & \parbox[t]{2mm}{\rotatebox[origin=c]{90}{CPS (our)}} & \parbox[t]{2mm}{\rotatebox[origin=c]{90}{Unpublished}} & \parbox[t]{2mm}{\rotatebox[origin=c]{90}{(AU)PRC}} & \parbox[t]{2mm}{\rotatebox[origin=c]{90}{(AU)ROC}} & \parbox[t]{2mm}{\rotatebox[origin=c]{90}{Accuracy}} & \parbox[t]{2mm}{\rotatebox[origin=c]{90}{Sensitivity}} & \parbox[t]{2mm}{\rotatebox[origin=c]{90}{Specificity}} & \parbox[t]{2mm}{\rotatebox[origin=c]{90}{Precision}} & \parbox[t]{2mm}{\rotatebox[origin=c]{90}{Recall}} & \parbox[t]{2mm}{\rotatebox[origin=c]{90}{$\beta$ for $F_\beta$}} & \parbox[t]{2mm}{\rotatebox[origin=c]{90}{Cohen's $\kappa$}} & \parbox[t]{2mm}{\rotatebox[origin=c]{90}{EEG}} & \parbox[t]{2mm}{\rotatebox[origin=c]{90}{EMG}} & \parbox[t]{2mm}{\rotatebox[origin=c]{90}{ECG}} &\parbox[t]{2mm}{\rotatebox[origin=c]{90}{Other}}\\
\cmidrule(lr){1-23}
\citet{badiei2023novel} & \textbf{x} & \textbf{x} & x & & x & x & & & & & x & x & x & & & & & & x & & x &\\
\citet{foroughi2023deep} & x & & x & & x & & & & & & x & x & x & x & & & & & x & & &\\
\citet{li2018sleep} & x & & x & & x & & & & & \textbf{x} & \textbf{x} & & & & & & & & x & x & &\\
\citet{kuo2023machine} & x & & x & & & & & & x & x & x & x & & & x & x & 1 & & & & & \textbf{x}\\
\citet{miller2018automatic} & x & & & \textbf{x} & x & & & & & \textbf{x} & \textbf{x} & & & & & & & & x & x & x & \textbf{x}\\
\citet{howe2018automated} & \textbf{x} & \textbf{x} & & \textbf{x} & x & & & & & \textbf{x} & \textbf{x} & & & & & & & & x & x & x & \textbf{x}\\
\citet{li2021deepsleep} & \textbf{x} & \textbf{x} & & \textbf{x} & x & x & & & & \textbf{x} & \textbf{x} & & & & & & & & x & x & x & \textbf{x}\\
\citet{fonod2022deepsleep} & x & & & \textbf{x} & x & & & & & \textbf{x} & \textbf{x} & & & & & & & & x & x & x & \textbf{x}\\
\citet{zan2023multi} & \textbf{x} & \textbf{x} & & \textbf{x} & & x & x & & & x & x & x & & & x & x & 1 & x & x & & &\\
our & x & & & \textbf{x} & x & & & \textbf{x} & & & & & & & x & x & \textbf{2} & & x & x & x & \textbf{x}\\

\bottomrule
\end{tabular}
\end{center}
\end{table}

\subsection{Methods for arousal detection}\label{sec:methods_arousal_detection}
Arousal detection often mirrors methods used in sleep stage classification, where the primary goal is to determine the sleep stage for each 30-second epoch of a polysomnographic recording. The data is typically segmented into $N$ consecutive windows of fixed length $s$, with $s$ either optimized as a hyperparameter or fixed at 30 seconds~\citep{li2018sleep, phan2019seqsleepnet} or other durations~\citep{kuo2023machine}. Overlapping windows are frequently employed to better capture arousal events, sometimes extended for evaluation as well~\citep{badiei2023novel, li2018sleep}. The definition of when a window signifies an arousal event often introduces another layer of complexity, sometimes determined by majority voting within the window or by the presence of at least one arousal label~\citep{kuo2023machine}, which may lead to another hyperparameter~\citep{li2018sleep}.
We build on the methodological foundation of the \textit{DeepSleep} architecture, which employs a Fully Convolutional Neural Network (FCN) with a U-Net architecture 
to process extensive polysomnographic signals continuously~\citep{li2021deepsleep}. This model differs from windowing approaches as it handles the entire dataset as a single sequence, where each point is evaluated within the context of its receptive field.
This comprehensive approach to arousal detection offers several advantages over traditional window-based methods: It eliminates the need for multiple hyperparameters, does not require manual feature extraction, supports an end-to-end process,
processes multimodal data natively, and leverages extensive temporal contexts to capture interactions across various timescales~\citep{li2021deepsleep}. This has spurred a growing body of research pursuing similar comprehensive methodologies, as documented in Table~\ref{tab:related_work}.

\subsection{Current state of evaluating arousal detection models}\label{sec:evaluation_arousal_detection}
Evaluating arousal detection models is challenging due to the diversity of methodologies employed, such as pointwise versus window-based evaluations, and the absence of standardized evaluation protocols~\citep{foroughi2023deep, badiei2023novel}. The wide array of performance metrics used further complicates the situation, as detailed in Table~\ref{tab:related_work}. Window-based evaluations are predominantly used over pointwise evaluations, reflecting the common practice in model training where window-based classification (WBC) is favored over continuous segmentation (CS) approaches.
For instance, \cite{zan2023multi} employ CS for both sleep stage classification and arousal detection in a multitask setup but still perform window-based evaluations alongside pointwise evaluations. They apply 30-second non-overlapping sliding windows on predictions, determining the window labels either by majority rule for sleep stages or the presence of any arousal indicator. We also employ both window-based and pointwise evaluations as baselines using a similar \textit{presence} criterion as \cite{zan2023multi}. However, we propose an approach closest to segment-wise f-score, introduced by \cite{hundman2018detecting} for time series anomaly detection (TSAD), which treats each contiguous segment of predictions as a singular event for comparison against true labeled sequences. We adapt this method to address certain limitations and align it more closely with the specific needs of arousal detection.
The term segment-wise f-score was coined by \cite{sorbo2023navigating}, who established a comprehensive taxonomy of evaluation measures for TSAD. This work criticizes the reliance on traditional evaluation metrics in TSAD and argues for adopting metrics tailored to the specific requirements of the task. Since arousal onset detection is closely related to supervised changepoint detection in TSAD, we will revisit the framework proposed by \citet{sorbo2023navigating} to embed our approach in their taxonomy.

\subsection{Datasets and data modalities for arousal detection}\label{sec:datasets}
Prominent datasets in sleep research include the 2018 PhysioNet Challenge dataset~\citep{ghassemi2018you, goldberger2000physiobank}, the Sleep Heart Health Study (SHHS)~\citep{quan1997sleep,zhang2018national}, 
and the Multi-Ethnic Study of Atherosclerosis (MESA)\citep{chen2015racial, zhang2018national}. Like our CPS dataset, they offer extensive PSG data and (in case of SHHS and MESA) patient information collected through standardized questionnaires such as sleep and restless legs questionnaires, the Pittsburgh Sleep Quality Index (PSQI), and the Epworth Sleepiness Scale (ESS).

However, beyond a broader set of raw PSG channels, our CPS dataset introduces several key advancements. It includes channels derived using the DOMINO expert software from SOMNOmedics GmbH, featuring innovative modalities like pulse transit time and beat-by-beat blood pressure estimations. The potential of these modalities in sleep diagnostics is supported by various studies~\citep{misaka2020clinical, argod1998differentiating, pitson1998value, pitson1994changes}, and we aim to further investigate their impact on arousal detection in an ongoing clinical study~\citep{wienhausen2024computer}.
Additional highlights of the CPS dataset include annotations indicating whether arousals were first detected in the EEG or as a consequence of other physiological changes, along with detailed medical outcomes such as sleep diagnoses, Baveno classification, and T90 value.
Further details on the CPS dataset are provided in Appendix~\ref{appendix:cps_dataset}.

\section{Methodologies}\label{sec:methods}
We first describe our main approach for arousal detection in Section~\ref{sec:training_continuous_segmentation}, present our novel framework ALPEC in Section~\ref{sec:evaluation_framework}, and then introduce multiple baseline approaches in Section~\ref{sec:baseline_approaches}.

\subsection{Arousal detection by continuous segmentation}\label{sec:training_continuous_segmentation}
For arousal onset detection, we adopt the DeepSleep architecture, which facilitates continuous segmentation of data into distinct classes~\citep{li2021deepsleep}. We build on an optimized version of this architecture, proposed by \citet{fonod2022deepsleep}, under MIT license, which reduces the U-Net's depth from 11 to 5 layers, substantially decreasing computational demands while maintaining comparable performance. This streamlined model processes all data points from multi-channel sleep recordings simultaneously, eliminating the need for window-based classification. It translates these inputs into sleep arousal scores for each data point using a binary cross-entropy (BCE) loss function. We refine this approach by employing a weighted BCE loss, adjusting the loss contribution of each data point by inversely weighting it according to the frequency of the arousal class within the subject's data, addressing class imbalance.
Since detecting singular arousal onset points does not work well with the DeepSleep approach (see Appendix~\ref{appendix:physionet_challenge}), we modify the ground-truth annotations to mark intervals of length $l$ around each arousal onset as positive. We select $l=10$ seconds which aligns with arousal scoring rules that require at least 10 seconds of stable sleep between distinct arousal events, ensuring the created ground-truth intervals do not overlap~\citep{berry2012rules}. We call this approach interval-based onset detection. During inference, the DeepSleep model outputs probability scores $p_i(\mathbf{x})$ for each data point $i$. To smooth these outputs for reducing false detections, we apply an averaging filter over a smoothing window of $w=3$ seconds on each point.

\subsection{ALPEC: Approximate localization and precise event count framework for post-processing and performance evaluation}\label{sec:evaluation_framework}
\citet{sorbo2023navigating} rightly state that there is no universally correct set of metrics for any specific task; however, using inappropriate metrics can lead to suboptimal decisions when selecting algorithms for productional use.
Guided by their taxonomy, we developed the Approximate Localization and Precise Event Count (ALPEC) framework to address the need for standardized evaluation metrics in arousal detection that align with the operational goals of clinical practice. Our framework is detailed in Algorithm~\ref{alg:alpec} and further formalized in Appendix~\ref{appendix:alpec}, with a schematic overview provided in Appendix~\ref{appendix:schemes_comparison}.

\begin{algorithm}[htbp]
\LinesNumbered
\small
\SetAlgoVlined
\SetKwInput{KwData}{Data}
\SetKwInput{KwIn}{Input}
\SetKwInput{KwOut}{Output}
\KwData{Multivariate input channels $\mathbf{x}$, training set $T$, validation set $V$, ground-truth intervals $G$}
\KwIn{Probability scores $p_i(\mathbf{x})$ or $p_{\eta}(\mathbf{x})$ for each data point $i$ or window $\eta$ or binary predictions $c_{\eta}$ for each window, and hyperparameters: Minimum interval merge distance $\delta$, maximum predicted interval duration $d$, ground-truth buffers $b^\text{before}$ and $b^\text{after}$}
\KwOut{Mean values for PRE (precision), REC (recall) and F2-score over subjects in $V$}

\SetKwFunction{FDetermineOptimalThreshold}{DetermineOptimalThresholdOnTrainingSet}
\SetKwFunction{FCompareTruthPredPerSubject}{CompareTruthPredPerSubject}
\SetKwFunction{FEval}{Eval}
\SetKwFunction{FPostProcessPredictions}{PostProcPreds}
\SetKwProg{Fn}{Function}{:}{\KwRet}

\BlankLine

\eIf{Input contains probability scores $p_i(\mathbf{x})$ or $p_{\eta}(\mathbf{x})$}{

$t_\text{opt} \gets$ \FDetermineOptimalThreshold{} \tcp*[r]{Get optimal threshold}\nllabel{line:get_opt_threshold}

}(\tcp*[h]Input contains binary predictions $c_{\eta}$){
$t_\text{opt} \gets None$ \tcp*[r]{No thresholding}
}

\ForEach{subject $\nu$ in $V$}{
    $\text{PRE}_\nu, \text{REC}_\nu, \text{F2}_\nu \gets$ \FEval{\FCompareTruthPredPerSubject{\FPostProcessPredictions{$\nu$, $t_\text{opt}$}}} \;\nllabel{line:eval_final_metrics}
}
$\bar{\text{PRE}}, \bar{\text{REC}}, \bar{\text{F2}} \gets$ Compute mean values over subjects $\nu$ \tcp*[r]{Get $\bar{\text{PRE}}, \bar{\text{REC}}, \bar{\text{F2}}$}\nllabel{line:calc_final_metrics}
\BlankLine

\Fn{\FDetermineOptimalThreshold{}}{

\ForEach{subject $\nu$ in $T$}{\nllabel{line:it_subjects}
\For{threshold $t=0, ..., 1$ in steps of $0.01$}{\nllabel{line:it_thresholds_train}

$c_{i\nu} \gets$ \FPostProcessPredictions{$\nu$, $t$} \tcp*[r]{Post-processing}\nllabel{line:post_processing}

            $\text{F2}_{\nu t} \gets$ \FEval{\FCompareTruthPredPerSubject{$c_{i\nu}$}} \tcp*[r]{Compare intervals}\nllabel{line:compare_intervals}
}
}
    $t_\text{opt} \gets$ Get threshold $t$ with the highest average $\text{F2}_{\nu t}$ over $\nu$ \tcp*[r]{Find optimal F2}\nllabel{line:update_opt_threshold}
    \KwRet $t_\text{opt}$ \tcp*[r]{Return optimal threshold}
}

\BlankLine

\Fn{\FPostProcessPredictions{$\nu$, $t$}}{

\eIf{window-based classification}{
Convert $p_{\eta\nu}(\mathbf{x})$ to binary predictions $c_{\eta\nu}$ if $t\neq None$ \tcp*[r]{Thresholding}\nllabel{line:thresholding_wbc}
Resample $c_{\eta\nu}$ to get binary predictions $c_{i\nu}$ per data point \tcp*[r]{Resampling}\nllabel{line:resampling}
    }(\tcp*[h]Continuous segmentation){
        Convert $p_{i\nu}(\mathbf{x})$ to binary predictions $c_{i\nu}$ using threshold $t$ \tcp*[r]{Thresholding}
    }\nllabel{line:thresholding_cs}
Merge intervals in $c_{i\nu}$ closer than $\delta$ \tcp*[r]{Interval merging}\nllabel{line:interval_merging}

\KwRet $c_{i\nu}$ \tcp*[r]{Return post-processed predictions}
}

\BlankLine

\Fn{\FCompareTruthPredPerSubject{$c_i$}}{
    Init TP, FP, FN to zero and empty set $M^\text{P}$ for tracking matched predicted intervals $P$ in $c_i$;\;
    
Extend each true interval $G$ by $b^\text{before}$ and $b^\text{after}$ \tcp*[r]{Buffer ground-truth}\nllabel{line:buffer_gt}

\ForEach{extended ground-truth interval $G$}{\nllabel{line:start_counting}
\eIf(\tcp*[f]{Selecting}){at least one overlap of $G$ with any $P \notin M^\text{P}$ exists with length($P$)$\le d$}{\nllabel{line:interval_filtering}
Add first overlapping $P$ to $M^\text{P}$\tcp*[r]{Track matched interval}
TP++ \tcp*[r]{Increment TP}
}{
FN++ \tcp*[r]{Increment FN}
}
}

Set FP to the number of predicted intervals $P$ not in $M^\text{P}$ \tcp*[r]{Count FP}\nllabel{line:end_counting}
\KwRet TP, FP, FN \tcp*[r]{Return TP, FP, FN}
}
\caption{\textbf{ALPEC post-processing and performance evaluation framework.} This compact representation assumes data and main input as globally accessible. The \textit{Eval} function is a placeholder for known implementations in the literature to calculate metrics from TP, FP and FN counts. For-loop variables used outside their scope imply storage in accumulative data structures.}\label{alg:alpec}
\end{algorithm}

\paragraph{Description of the ALPEC procedure for arousal detection}
ALPEC is compatible with both window-based classification (WBC) and continuous segmentation (CS) approaches to arousal detection. For WBC, we process either probability scores $p_{\eta}(\mathbf{x})$ or binary predictions $c_{\eta}$ for each window $\eta$. For CS, the process begins with probability scores $p_{i}(\mathbf{x})$ for each data point $i$. In cases where probability scores are available, the optimal threshold $t_\text{opt}$ is determined from the training set $T$ to convert these scores into binary predictions (Algorithm~\ref{alg:alpec}, line \ref{line:get_opt_threshold}).
For each subject $\nu$ and threshold $t$, we post-process the predictions by applying thresholding and resampling for WBC (lines \ref{line:thresholding_wbc} and \ref{line:resampling}) or just thresholding for CS (line \ref{line:thresholding_cs}). 
Next, predicted intervals that are less than 10 seconds apart are merged (line \ref{line:interval_merging}). For full event detection, merging is based on the closest points of predicted intervals, while for arousal onset detection, it is based on the maxima of the prediction scores, indicating the most likely points of arousal onset.

After post-processing, predictions are compared to ground-truth data $G$ -- which may consist of full event annotations, point annotations, or constructed intervals (see Section~\ref{sec:training_continuous_segmentation}) -- to determine true positive (TP), false positive (FP), and false negative (FN) counts (line \ref{line:compare_intervals}).

ALPEC introduces two key \textit{approximate localization} components. First, a temporal tolerance buffer~\citep{scharwachter2020statistical} of 15 seconds is applied before and after each ground-truth interval (line \ref{line:buffer_gt}). Predicted intervals overlapping with ground-truth intervals within this buffer are counted as TPs. Second, ALPEC restricts the maximum duration of predicted intervals, with only those shorter than 60 seconds qualifying as TPs (line \ref{line:interval_filtering}).

The counting method in ALPEC (lines \ref{line:start_counting} to \ref{line:end_counting}) fulfills the \textit{precise event count} requirement. A TP is recorded when any eligible predicted interval $P$ overlaps with a buffered ground-truth interval $G$. A FN is recorded if a $G$ does not overlap with any $P$, and a FP is noted if a $P$ does not overlap with any $G$. Overly long predicted intervals contribute to only one TP, and multiple ground-truth intervals spanned by a single predicted interval result in multiple FNs unless each ground-truth interval is uniquely matched to a predicted interval.

Selecting appropriate metrics is crucial for evaluating and analyzing model performance. Following the taxonomy by \citet{sorbo2023navigating}, we select the F2 score as the final metric for optimization (performance evaluation) and use precision and recall as auxiliary metrics for additional insights (performance analysis). 
ALPEC computes the micro-average F2 scores across all subjects in the training set $T$ to determine the optimal decision threshold $t_\text{opt}$ (line \ref{line:update_opt_threshold}). This threshold is then used to calculate the metrics for each subject in the validation set $V$, which may also be the test set (line \ref{line:eval_final_metrics}). Results are aggregated using the mean to adequately represent individual outliers (line \ref{line:calc_final_metrics}).

\paragraph{Rationale and comparison to existing metrics}

ALPEC shares several similarities with existing metrics. First, it is most similar to the segment-wise f-score~\citep{hundman2018detecting}, as both approaches focus on evaluating segment overlaps rather than pointwise predictions. 
Second, like some existing methods, ALPEC employs a temporal tolerance buffer around ground-truth intervals~\citep{scharwachter2020statistical}, corresponding to the typical length of one 30-second epoch as viewed by medical scorers. The use of this buffer addresses potential temporal inaccuracies in arousal event annotations and is backed by the irrelevance of precise annotations in current clinical practice, making the evaluation process more robust and clinically relevant.
Third, the integration of ALPEC into the taxonomy by \citet{sorbo2023navigating} demonstrates its alignment with established categories in time series anomaly detection. ALPEC metrics classify as \textit{binary}, since thresholding does not manipulate prediction scores, and \textit{redefined counting-based}, as they involve comparing intervals rather than evaluating pointwise. Additionally, they exhibit intrinsic insensitivity to true negatives and a valuation property of time tolerance.

ALPEC also introduces several significant differences:
First, ALPEC merges close predicted intervals, in line with clinical guidelines requiring at least 10 seconds of stable sleep between arousals~\citep{berry2012rules}.
Second, ALPEC imposes a maximum duration on predicted intervals, ensuring they do not exceed practical lengths. This restriction prevents ambiguities during human review and maintains the clinical relevance of arousal timing relative to sleep stages which is important for physicians in sleep medicine.
Third, the method of precise event counting ensures that only one TP is counted per predicted interval, which is essential for clinical utility. If a predicted interval spans multiple ground-truth intervals, it results in multiple FNs unless each ground-truth interval is uniquely matched to a predicted interval. This approach avoids the pitfall of inaccurately rewarding temporal extension of predicted intervals, a limitation of segment-wise f-score methods~\citep{sorbo2023navigating}.
Finally, unlike most current approaches to arousal detection, which typically rely on the F1 score or limit their reports to AUPRC or AUROC without a clear consensus on the most appropriate metric (see Table \ref{tab:related_work}), ALPEC follows \citet{sorbo2023navigating} in advocating for a context-aware selection of metrics tailored to the operational environment.
AI-based clinical decision support systems (CDSS) in healthcare aim not only to improve the quality of outcomes but also to enhance the efficiency of medical practitioners' work~\citep{magrabi2019artificial, vasey2022reporting}. Given the need to relieve medical practitioners from reviewing extensive amounts of data -- in our case night-long recordings -- it is crucial for CDSS to highlight the most pertinent data sections.
This objective is best met by ensuring that AI predictions minimize missed arousals (false negatives), allowing human reviewers to efficiently address any false positives. The F2 score is particularly well-suited for this purpose, as it prioritizes recall over precision, reducing the likelihood of overlooked arousal events.
While AUPRC is favored by some for its perceived robustness against class imbalance \citep{qian2021review}, recent critiques highlight its potential biases, indicating the need for a more nuanced approach~\citep{mcdermott2024closer}. More critically, relying solely on AUPRC or AUROC is inadequate for clinical decision-making, as these metrics do not fully capture how model predictions translate into actionable clinical outcomes. Therefore, we acknowledge their utility in performance analysis but do not consider them sufficient for performance evaluation.

\subsection{Baseline approaches}\label{sec:baseline_approaches}

In this work, 
we also explore traditional window-based classification (WBC) methods. These WBC approaches are evaluated using both standard window-based evaluation and our ALPEC framework. We employ several classical univariate models from the sktime library~\citep{loning2019sktime} (BSD 3-Clause License), utilizing them with their standard configurations. 
For each arousal onset in the training set, we construct a 30-second window centered randomly around the onset point to enhance generalizability. This random alignment aims to mimic the variable alignment of arousal onset points during inference across non-overlapping windows covering the entire series of a subject.
To address the challenge of class imbalance, we select an equal number of negative-class windows randomly for each subject during training. The evaluation then proceeds with standard WBC, dividing the test set data into consecutive non-overlapping windows of the same 30-second length used in training.
Adopting window-based onset detection simplifies the windowing approach by reducing the need for overlapping windows or complex voting schemes. Instead, we apply the \textit{presence} criterion to determine the class of the windows, aiming to approximate the original class distribution in the test set.
Additionally, we ensure robustness by conducting cross-subject validation, where training and testing sets include distinct subjects.
Furthermore, we conduct baseline experiments where CS approaches are evaluated using both traditional window-based evaluation and ALPEC for comparative analysis. This involves creating windows for ground-truth and prediction as described above. 

\section{Results}\label{sec:main_results}
We first introduce our CPS dataset in Section~\ref{sec:cps_dataset}. Then, in Section~\ref{sec:comparison_evaluation_schemes}, we compare window-based classification (WBC) and continuous segmentation (CS) training approaches for arousal onset detection, evaluating them through both window-based evaluation and ALPEC, assessing the utility of different data modalities. Additional comparative analyses of training and evaluation schemes on another dataset and ablation studies on hyperparameter choices are detailed in Appendices~\ref{appendix:physionet_challenge} and~\ref{appendix:ablation_studies}.

\subsection{Comprehensive polysomnography (CPS) dataset}\label{sec:cps_dataset}
We collected the Comprehensive Polysomnography (CPS) dataset during clinical practice from 2021-2022 in a state-of-the-art sleep laboratory at Klinikum Esslingen, Germany. It is released alongside this work on the PhysioNet platform~\citep{kraft2024cps, goldberger2000physiobank}. The dataset comprises 113 diagnostic polysomnographic recordings, encompassing up to 36 raw and 23 derived data channels, alongside 81 types of annotated events and additional questionnaire data for each subject. The dataset annotates various arousal classes, including those related to respiratory efforts, flow limitations, oxygen desaturation, limb movements, and spontaneous arousals. We combine all classes into a single category for binary event detection. Further details on the CPS dataset are available in Appendix~\ref{appendix:cps_dataset}.

\subsection{Comparison of training and evaluation schemes}\label{sec:comparison_evaluation_schemes}
Using the CPS dataset, we trained DeepSleep model candidates D1-D4, which utilize continuous segmentation on interval-based onset detection (IOD, see Section \ref{sec:training_continuous_segmentation}). They are compared to popular time series classification models and naive baselines from the sktime library~\citep{loning2019sktime} on window-based onset detection (WOD, see Section \ref{sec:baseline_approaches}).
All models are trained using combined training and validation folds (93 subjects total). The DeepSleep models are trained until early stopping or up to 100 epochs. All models are evaluated on the held-out test set (14 subjects) using both window-based evaluation (WE) and ALPEC. All experiments are repeated five times using different random seeds. Details on data preprocessing, hyperparameter choices (including selected channels for models D3 and D4), and data folds are available in Appendices~\ref{appendix:data_preprocessing},~\ref{appendix:hyperparameter_tuning}, and~\ref{appendix:data_folds}. The results are shown in Table~\ref{tab:comparion_evaluation_schemes}.

\begin{table}[htbp]
\caption{\textbf{Comparative evaluation of modeling approaches using ALPEC and window-based evaluation (WE).} Performance metrics are presented as mean values over distinct test subjects with 95\% confidence intervals in brackets, assuming t-distributed mean values, calculated over five training iterations. For window-based onset detection  (WOD) and baseline models, methods from the sktime library~\citep{loning2019sktime}, including dummy models, were used. Models D1-D4, derived from the DeepSleep approach (see Section~\ref{sec:training_continuous_segmentation}), indicate the channels utilized. ALPEC offers a stringent assessment of performance, highlighting that WE tends to overestimate the effectiveness of WOD models.}\label{tab:comparion_evaluation_schemes}
\begin{center}
\setlength{\tabcolsep}{3.8pt} %
\begin{tabular}{llllllll}
\toprule
& \multirow{2}{*}{Model} & \multicolumn{3}{c}{Window-based evaluation (WE)} & \multicolumn{3}{c}{ALPEC evaluation}\\
\cmidrule(lr){3-5}
\cmidrule(lr){6-8}
& & $\bar{\text{Precision}}$ & $\bar{\text{Recall}}$ & $\bar{\text{F}}2$ & $\bar{\text{Precision}}$ & $\bar{\text{Recall}}$ & $\bar{\text{F}}2$ \\
\midrule
\parbox[t]{2mm}{\multirow{4}{*}{\rotatebox[origin=c]{90}{IOD}}}

& D4: most channels & \errfmt{0.49}{6} & \errfmt{0.82}{8} & \errfmt{\textbf{0.71}}{3} & \errfmt{0.59}{8} & \errfmt{0.81}{8} & \errfmt{\textbf{0.73}}{3} \\

& D3: no EEG, EOG, EMG & \errfmt{0.39}{3} & \errfmt{0.71}{3} & \errfmt{0.59}{3} & \errfmt{0.48}{4} & \errfmt{0.70}{3} & \errfmt{0.62}{3} \\

& D2: C3:A2, EOGl, EMG & \errfmt{0.44}{5} & \errfmt{0.75}{7} & \errfmt{0.64}{3} & \errfmt{0.53}{7} & \errfmt{0.74}{7} & \errfmt{0.67}{3} \\

& D1: C3:A2 & \errfmt{0.40}{5} & \errfmt{0.76}{6} & \errfmt{0.62}{2} & \errfmt{0.48}{7} & \errfmt{0.75}{6} & \errfmt{0.65}{2} \\
\midrule
\parbox[t]{2mm}{\multirow{6}{*}{\rotatebox[origin=c]{90}{WOD}}}
& IndividualBOSS & \errfmt{0.25}{0} & \errfmt{0.55}{1} & \errfmt{0.42}{1} & \errfmt{0.30}{1} & \errfmt{0.32}{2} & \errfmt{0.31}{2}\\

& SupervisedTimeSeriesForest & \errfmt{0.37}{0} & \errfmt{0.71}{1} & \errfmt{0.59}{1} & \errfmt{0.37}{1} & \errfmt{0.29}{1} & \errfmt{0.30}{1}\\

& TimeSeriesForestClassifier & \errfmt{0.30}{1} & \errfmt{0.65}{1} & \errfmt{0.50}{1} & \errfmt{0.30}{1} & \errfmt{0.30}{1} & \errfmt{0.29}{1}\\

& SignatureClassifier & \errfmt{0.28}{0} & \errfmt{0.65}{2} & \errfmt{0.49}{1} & \errfmt{0.29}{1} & \errfmt{0.30}{1} & \errfmt{0.29}{1}\\

& SummaryClassifier & \errfmt{0.27}{0} & \errfmt{0.62}{1} & \errfmt{0.48}{0} & \errfmt{0.27}{1} & \errfmt{0.29}{1} & \errfmt{0.28}{1}\\

& Catch22Classifier & \errfmt{0.33}{0} & \errfmt{0.73}{1} & \errfmt{0.57}{0} & \errfmt{0.30}{0} & \errfmt{0.23}{1} & \errfmt{0.24}{1}\\

\midrule
\parbox[t]{2mm}{\multirow{4}{*}{\rotatebox[origin=c]{90}{Baseline}}}

& RandomStratified & \errfmt{0.20}{1} & \errfmt{0.49}{2} & \errfmt{0.37}{1} & \errfmt{0.29}{2} & \errfmt{0.36}{4} & \errfmt{0.34}{3}\\
& RandomUniform & \errfmt{0.20}{0} & \errfmt{0.51}{2} & \errfmt{0.37}{1} & \errfmt{0.28}{2} & \errfmt{0.36}{3} & \errfmt{0.33}{3}\\
& Constant 1 & 0.20 & 1.00 & 0.53 & 0.00 & 0.00 & 0.00\\
& Constant 0 & 0.00 & 0.00 & 0.00 & 0.00 & 0.00 & 0.00\\
\bottomrule
\end{tabular}
\end{center}
\end{table}

Both Window-Based Evaluation (WE) and ALPEC demonstrate similar performance across our interval-based onset (IOD) detection models (D1-D4), which utilize the DeepSleep approach. This consistency suggests that the domain-specific adaptations inherent in ALPEC do not drastically alter results compared to WE in this context. However, WE appears to underestimate precision, resulting in a higher count of false positives. This discrepancy is due to ALPEC's methodology of treating all adjacent points within an overlapping interval as a single true positive, unlike WE, which evaluates each window individually. Furthermore, ALPEC's buffer zones typically reduce false positives and negatives, enhancing its accuracy.
In the analysis of Window-Based Onset Detection (WOD) models, WE notably overestimates recall compared to ALPEC. A clear example is observed with the Constant 1 baseline, where WE significantly overrates its performance because this model predicts an arousal event in every window. This outcome reveals a bias in WE towards models that predict frequent events. Conversely, ALPEC shows zero performance for this baseline, effectively highlighting its ability to address the methodological shortcomings of its closest predecessor, the segment-wise f-score, as discussed in Section~\ref{sec:evaluation_framework}.

Overall, ALPEC provides a more accurate assessment of model performance, indicating that none of the WOD models significantly surpass the random baselines. This outcome is consistent with the finding that arousal detection, which is considered a challenging task requiring specialized feature engineering when not using advanced Deep Learning techniques, may not be effectively addressed by comparativley simple models~\citep{zan2023multi}.

\section{Discussion}\label{sec:discussion}

Our results demonstrate that arousal onset detection can be effectively achieved using continuous segmentation approaches with our proposed interval-based onset detection (IOD) training scheme. They also highlight the potential of using novel data modalities for enhanced predictive performance (model D4 in Table~\ref{tab:comparion_evaluation_schemes}) or reduced technical complexity (model D3). Minimizing dependence on electrode-based modalities could address issues such as electrode displacement or noise, potentially enabling home-based arousal diagnostics~\citep{imtiaz2021systematic}.
A significant contribution of our work is the development of the ALPEC framework, the first performance evaluation framework tailored to the clinical requirements of arousal detection. We demonstrate that ALPEC provides a more accurate assessment of model performance compared to traditional window-based evaluation (WE). Moreover, due to sampling at the subject-level, ALPEC overcomes common pitfalls of window-based evaluation such as class imbalance and cross-subject validation issues (see Appendix~\ref{appendix:evaluation_pitfalls} for further details).
We emphasize that our critique is not directed against window-based classification approaches, which remain valuable and effective in arousal detection, as shown in recent studies~\citep{badiei2023novel, foroughi2023deep}. Our concerns specifically relate to window-based evaluation methods. We advocate for the adoption of the ALPEC framework, which is immune to common pitfalls, finely tunable, and compatible with both window-based classification and continuous segmentation.
ALPEC's design incorporates a built-in \textit{precise event count} requirement, while also offering flexibility in \textit{approximate location} through adjustable parameters for buffer size and maximum interval length. This adaptability makes ALPEC a versatile tool for tasks that require precise event count detection in time series data, providing a robust framework that can accommodate a wide range of applications in healthcare and beyond.

\subsection{Limitations}\label{sec:limitations}
This work addresses binary detection of arousal onsets and does not encompass the causal differentiation of arousals, an additional task in clinical settings that necessitates a multi-class classification approach.
Furthermore, the ALPEC framework is designed solely for post-processing and performance evaluation, and does not influence the learning process of the models. While we adapted the DeepSleep method for arousal onset detection, it was not originally crafted with the specific requirements of real-world clinical tasks in mind. Future research should not only leverage ALPEC for comparative analysis but also enhance model functionality by incorporating factors crucial for clinical decision support, such as explainability.
Additionally, the settings for ALPEC's buffer size and maximum interval length hyperparameters need experimental tuning and validation in collaboration with clinical end-users to ensure their effectiveness and applicability in real-world settings.

\subsection{Conclusion}\label{sec:conclusion}
Our work establishes foundational elements for developing clinical decision support systems (CDSS) for arousal detection in sleep laboratories, addressing critical misalignments between current Machine Learning methodologies and clinical practices. We introduce the Comprehensive Polysomnography (CPS) dataset as a significant resource for sleep medical research, demonstrating the potential of utilizing novel data modalities.
Our findings contribute to the development of production-ready arousal detection models that align with current clinical annotation practices.
We look forward to seeing how the research community builds on our findings and continues to evolve the field.

\begin{ack}

This research was funded by STZ Softwaretechnik GmbH, a member of IT-Designers Gruppe, located in Esslingen am Neckar, Germany. 
We would like to extend our special thanks to Eduard Gindullis for his work on quality control of data collection and documentation, Dr. Rolf Wagner for his contribution to the clinical study design, Aurelia Mehl-Jöbstl and Dr. Andreas Rau for their guidance in matters of data protection and security, Alexander Harm and Tim Bauer for their assistance with coding, Tom Joseph Pollard and the team behind PhysioNet for their great service, and Claus-Dieter Weiss and the staff from NRI Medizintechnik GmbH for their cooperation in collecting the data and conducting the clinical study.
\end{ack}

\medskip

{\small
\bibliographystyle{plainnat}
\bibliography{sample}

\begin{thebibliography}{50}
\providecommand{\natexlab}[1]{#1}
\providecommand{\url}[1]{\texttt{#1}}
\expandafter\ifx\csname urlstyle\endcsname\relax
  \providecommand{\doi}[1]{doi: #1}\else
  \providecommand{\doi}{doi: \begingroup \urlstyle{rm}\Url}\fi

\bibitem[Akhtar et~al.(2024)Akhtar, Benjelloun, Conforti, Giner-Miguelez, Jain,
  Kuchnik, Lhoest, Marcenac, Maskey, Mattson, Oala, Ruyssen, Shinde, Simperl,
  Thomas, Tykhonov, Vanschoren, Vogler, and Wu]{akhtar2024croissant}
Mubashara Akhtar, Omar Benjelloun, Costanza Conforti, Joan Giner-Miguelez,
  Nitisha Jain, Michael Kuchnik, Quentin Lhoest, Pierre Marcenac, Manil Maskey,
  Peter Mattson, Luis Oala, Pierre Ruyssen, Rajat Shinde, Elena Simperl,
  Goeffry Thomas, Slava Tykhonov, Joaquin Vanschoren, Steffen Vogler, and
  Carole-Jean Wu.
\newblock {Croissant: A Metadata Format for ML-Ready Datasets}, 2024.

\bibitem[Aminikhanghahi and Cook(2017)]{aminikhanghahi2017survey}
Samaneh Aminikhanghahi and Diane~J Cook.
\newblock {A survey of methods for time series change point detection}.
\newblock \emph{Knowledge and information systems}, 51\penalty0 (2):\penalty0
  339--367, 2017.

\bibitem[Anido-Alonso and Alvarez-Estevez(2023)]{anido2023decentralized}
Adriana Anido-Alonso and Diego Alvarez-Estevez.
\newblock {Decentralized data-privacy preserving deep-learning approaches for
  enhancing inter-database generalization in automatic sleep staging}.
\newblock \emph{IEEE Journal of Biomedical and Health Informatics}, 2023.

\bibitem[Argod et~al.(1998)Argod, Pepin, and Levy]{argod1998differentiating}
Jerome Argod, Jean-Louis Pepin, and Patrick Levy.
\newblock Differentiating obstructive and central sleep respiratory events
  through pulse transit time.
\newblock \emph{American journal of respiratory and critical care medicine},
  158\penalty0 (6):\penalty0 1778--1783, 1998.

\bibitem[Badiei et~al.(2023)Badiei, Meshgini, and Rezaee]{badiei2023novel}
Afsoon Badiei, Saeed Meshgini, and Khosro Rezaee.
\newblock {A novel approach for sleep arousal disorder detection based on the
  interaction of physiological signals and metaheuristic learning}.
\newblock \emph{Computational Intelligence and Neuroscience}, 2023, 2023.

\bibitem[Berry et~al.(2012)Berry, Budhiraja, Gottlieb, Gozal, Iber, Kapur,
  Marcus, Mehra, Parthasarathy, Quan, et~al.]{berry2012rules}
Richard~B Berry, Rohit Budhiraja, Daniel~J Gottlieb, David Gozal, Conrad Iber,
  Vishesh~K Kapur, Carole~L Marcus, Reena Mehra, Sairam Parthasarathy, Stuart~F
  Quan, et~al.
\newblock {Rules for scoring respiratory events in sleep: update of the 2007
  AASM manual for the scoring of sleep and associated events: deliberations of
  the sleep apnea definitions task force of the American Academy of Sleep
  Medicine}.
\newblock \emph{Journal of clinical sleep medicine}, 8\penalty0 (5):\penalty0
  597--619, 2012.

\bibitem[Bertrand(2020)]{bertrand2020sweetviz}
Bertrand.
\newblock {SweetViz. Visualize and compare datasets, target values and
  associations, with one line of code.}
\newblock \url{https://github.com/fbdesignpro/sweetviz}, 2020.
\newblock Accessed: 2024-06-04.

\bibitem[Bonsignore et~al.(2019)Bonsignore, Saaresranta, and
  Riha]{bonsignore2019sex}
Maria~R Bonsignore, Tarja Saaresranta, and Renata~L Riha.
\newblock {Sex differences in obstructive sleep apnoea}.
\newblock \emph{European Respiratory Review}, 28\penalty0 (154), 2019.

\bibitem[Chen et~al.(2015)Chen, Wang, Zee, Lutsey, Javaheri, Alc{\'a}ntara,
  Jackson, Williams, and Redline]{chen2015racial}
Xiaoli Chen, Rui Wang, Phyllis Zee, Pamela~L Lutsey, Sogol Javaheri, Carmela
  Alc{\'a}ntara, Chandra~L Jackson, Michelle~A Williams, and Susan Redline.
\newblock {Racial/ethnic differences in sleep disturbances: the Multi-Ethnic
  Study of Atherosclerosis (MESA)}.
\newblock \emph{Sleep}, 38\penalty0 (6):\penalty0 877--888, 2015.

\bibitem[Fawzy et~al.(2023)Fawzy, Nirmala, Khansa, and
  Wardhana]{fawzy2023ethics}
Ahmad Fawzy, Danastri~Cantya Nirmala, Denaya Khansa, and Yudhistira~Tri
  Wardhana.
\newblock {Ethics and Regulation for Artificial Intelligence in Healthcare:
  Empowering Clinicians to Ensure Equitable and High-Quality Care}.
\newblock 2023.

\bibitem[Fietze et~al.(2019)Fietze, Laharnar, Obst, Ewert, Felix, Garcia,
  Gl{\"a}ser, Glos, Schmidt, Stubbe, et~al.]{fietze2019prevalence}
Ingo Fietze, Naima Laharnar, Anne Obst, Ralf Ewert, Stephan~B Felix, Carmen
  Garcia, Sven Gl{\"a}ser, Martin Glos, Carsten~Oliver Schmidt, Beate Stubbe,
  et~al.
\newblock {Prevalence and association analysis of obstructive sleep apnea with
  gender and age differences--Results of SHIP-Trend}.
\newblock \emph{Journal of sleep research}, 28\penalty0 (5):\penalty0 e12770,
  2019.

\bibitem[Fonod(2022)]{fonod2022deepsleep}
Robert Fonod.
\newblock {DeepSleep 2.0: automated sleep arousal segmentation via deep
  learning}.
\newblock \emph{AI}, 3\penalty0 (1):\penalty0 164--179, 2022.

\bibitem[Foroughi et~al.(2023)Foroughi, Farokhi, Rahatabad, and
  Kashaninia]{foroughi2023deep}
Andia Foroughi, Fardad Farokhi, Fereidoun~Nowshiravan Rahatabad, and Alireza
  Kashaninia.
\newblock {Deep convolutional architecture-based hybrid learning for sleep
  arousal events detection through single-lead EEG signals}.
\newblock \emph{Brain and Behavior}, 13\penalty0 (6):\penalty0 e3028, 2023.

\bibitem[Franklin and Lindberg(2015)]{franklin2015obstructive}
Karl~A Franklin and Eva Lindberg.
\newblock {Obstructive sleep apnea is a common disorder in the population—a
  review on the epidemiology of sleep apnea}.
\newblock \emph{Journal of thoracic disease}, 7\penalty0 (8):\penalty0 1311,
  2015.

\bibitem[Garza and Mergenthaler-Canseco(2023)]{garza2023timegpt}
Azul Garza and Max Mergenthaler-Canseco.
\newblock {TimeGPT-1}.
\newblock \emph{arXiv preprint arXiv:2310.03589}, 2023.

\bibitem[Gebru et~al.(2021)Gebru, Morgenstern, Vecchione, Vaughan, Wallach,
  Iii, and Crawford]{gebru2021datasheets}
Timnit Gebru, Jamie Morgenstern, Briana Vecchione, Jennifer~Wortman Vaughan,
  Hanna Wallach, Hal~Daum{\'e} Iii, and Kate Crawford.
\newblock {Datasheets for datasets}.
\newblock \emph{Communications of the ACM}, 64\penalty0 (12):\penalty0 86--92,
  2021.

\bibitem[Ghassemi et~al.(2018)Ghassemi, Moody, Lehman, Song, Li, Sun, Mark,
  Westover, and Clifford]{ghassemi2018you}
Mohammad~M Ghassemi, Benjamin~E Moody, Li-Wei~H Lehman, Christopher Song, Qiao
  Li, Haoqi Sun, Roger~G Mark, M~Brandon Westover, and Gari~D Clifford.
\newblock {You snooze, you win: the physionet/computing in cardiology challenge
  2018}.
\newblock In \emph{{2018 Computing in Cardiology Conference (CinC)}},
  volume~45, pages 1--4. IEEE, 2018.

\bibitem[Goldberger et~al.(2000)Goldberger, Amaral, Glass, Hausdorff, Ivanov,
  Mark, Mietus, Moody, Peng, and Stanley]{goldberger2000physiobank}
Ary~L Goldberger, Luis~AN Amaral, Leon Glass, Jeffrey~M Hausdorff, Plamen~Ch
  Ivanov, Roger~G Mark, Joseph~E Mietus, George~B Moody, Chung-Kang Peng, and
  H~Eugene Stanley.
\newblock {PhysioBank, PhysioToolkit, and PhysioNet: components of a new
  research resource for complex physiologic signals}.
\newblock \emph{circulation}, 101\penalty0 (23):\penalty0 e215--e220, 2000.

\bibitem[Howe-Patterson et~al.(2018)Howe-Patterson, Pourbabaee, and
  Benard]{howe2018automated}
Matthew Howe-Patterson, Bahareh Pourbabaee, and Frederic Benard.
\newblock {Automated detection of sleep arousals from polysomnography data
  using a dense convolutional neural network}.
\newblock In \emph{{2018 Computing in Cardiology Conference (CinC)}},
  volume~45, pages 1--4. IEEE, 2018.

\bibitem[Hundman et~al.(2018)Hundman, Constantinou, Laporte, Colwell, and
  Soderstrom]{hundman2018detecting}
Kyle Hundman, Valentino Constantinou, Christopher Laporte, Ian Colwell, and Tom
  Soderstrom.
\newblock {Detecting spacecraft anomalies using lstms and nonparametric dynamic
  thresholding}.
\newblock In \emph{{Proceedings of the 24th ACM SIGKDD international conference
  on knowledge discovery \& data mining}}, pages 387--395, 2018.

\bibitem[Imtiaz(2021)]{imtiaz2021systematic}
Syed~Anas Imtiaz.
\newblock {A systematic review of sensing technologies for wearable sleep
  staging}.
\newblock \emph{Sensors}, 21\penalty0 (5):\penalty0 1562, 2021.

\bibitem[Jehan et~al.(2017)Jehan, Zizi, Pandi-Perumal, Wall, Auguste, Myers,
  Jean-Louis, and McFarlane]{jehan2017obstructive}
Shazia Jehan, Ferdinand Zizi, Seithikurippu~R Pandi-Perumal, Steven Wall, Evan
  Auguste, Alyson~K Myers, Girardin Jean-Louis, and Samy~I McFarlane.
\newblock {Obstructive sleep apnea and obesity: implications for public
  health}.
\newblock \emph{Sleep medicine and disorders: international journal},
  1\penalty0 (4), 2017.

\bibitem[Kraft et~al.(2024)Kraft, Theissler, Wienhausen-Wilke, Walter, and
  Kasneci]{kraft2024cps}
Stefan Kraft, Andreas Theissler, Vera Wienhausen-Wilke, Philipp Walter, and
  Gjergji Kasneci.
\newblock {Comprehensive Polysomnography (CPS) Dataset: A Resource for
  Sleep-Related Arousal Research (version 1.0.0)}.
\newblock PhysioNet, 2024.
\newblock URL \url{https://doi.org/10.13026/sxs0-h317}.

\bibitem[Kuo et~al.(2023)Kuo, Tsai, Cheng, Hs, Majumdar, Stettler, Lee, Kuan,
  Feng, Tseng, et~al.]{kuo2023machine}
Chih-Fan Kuo, Cheng-Yu Tsai, Wun-Hao Cheng, Wen-Hua Hs, Arnab Majumdar, Marc
  Stettler, Kang-Yun Lee, Yi-Chun Kuan, Po-Hao Feng, Chien-Hua Tseng, et~al.
\newblock {Machine learning approaches for predicting sleep arousal response
  based on heart rate variability, oxygen saturation, and body profiles}.
\newblock \emph{Digital Health}, 9:\penalty0 20552076231205744, 2023.

\bibitem[Lacoste et~al.(2019)Lacoste, Luccioni, Schmidt, and
  Dandres]{lacoste2019quantifying}
Alexandre Lacoste, Alexandra Luccioni, Victor Schmidt, and Thomas Dandres.
\newblock {Quantifying the Carbon Emissions of Machine Learning}.
\newblock \emph{arXiv preprint arXiv:1910.09700}, 2019.

\bibitem[Li et~al.(2018)Li, Cao, Zhong, and Pan]{li2018sleep}
Haoqi Li, Qineng Cao, Yizhou Zhong, and Yun Pan.
\newblock {Sleep arousal detection using end-to-end deep learning method based
  on multi-physiological signals}.
\newblock In \emph{{2018 computing in cardiology conference (CinC)}},
  volume~45, pages 1--4. IEEE, 2018.

\bibitem[Li and Guan(2021)]{li2021deepsleep}
Hongyang Li and Yuanfang Guan.
\newblock {DeepSleep convolutional neural network allows accurate and fast
  detection of sleep arousal}.
\newblock \emph{Communications biology}, 4\penalty0 (1):\penalty0 18, 2021.

\bibitem[L{\"o}ning et~al.(2019)L{\"o}ning, Bagnall, Ganesh, Kazakov, Lines,
  and Kir{\'a}ly]{loning2019sktime}
Markus L{\"o}ning, Anthony Bagnall, Sajaysurya Ganesh, Viktor Kazakov, Jason
  Lines, and Franz~J Kir{\'a}ly.
\newblock {sktime: A unified interface for machine learning with time series}.
\newblock \emph{arXiv preprint arXiv:1909.07872}, 2019.

\bibitem[Madiega(2021)]{madiega2021artificial}
Tambiama Madiega.
\newblock {Artificial intelligence act}.
\newblock \emph{European Parliament: European Parliamentary Research Service},
  2021.

\bibitem[Magrabi et~al.(2019)Magrabi, Ammenwerth, McNair, De~Keizer,
  Hypp{\"o}nen, Nyk{\"a}nen, Rigby, Scott, Vehko, Wong,
  et~al.]{magrabi2019artificial}
Farah Magrabi, Elske Ammenwerth, Jytte~Brender McNair, Nicolet~F De~Keizer,
  Hannele Hypp{\"o}nen, Pirkko Nyk{\"a}nen, Michael Rigby, Philip~J Scott,
  Tuulikki Vehko, Zoie Shui-Yee Wong, et~al.
\newblock {Artificial intelligence in clinical decision support: challenges for
  evaluating AI and practical implications}.
\newblock \emph{Yearbook of medical informatics}, 28\penalty0 (01):\penalty0
  128--134, 2019.

\bibitem[McDermott et~al.(2024)McDermott, Hansen, Zhang, Angelotti, and
  Gallifant]{mcdermott2024closer}
Matthew McDermott, Lasse~Hyldig Hansen, Haoran Zhang, Giovanni Angelotti, and
  Jack Gallifant.
\newblock {A Closer Look at AUROC and AUPRC under Class Imbalance}.
\newblock \emph{arXiv preprint arXiv:2401.06091}, 2024.

\bibitem[Miller et~al.(2018)Miller, Ward, and Bambos]{miller2018automatic}
Daniel Miller, Andrew Ward, and Nicholas Bambos.
\newblock {Automatic sleep arousal identification from physiological waveforms
  using deep learning}.
\newblock In \emph{{2018 Computing in Cardiology Conference (CinC)}},
  volume~45, pages 1--4. IEEE, 2018.

\bibitem[Misaka et~al.(2020)Misaka, Niimura, Yoshihisa, Wada, Kimishima,
  Yokokawa, Abe, Oikawa, Kaneshiro, Kobayashi, et~al.]{misaka2020clinical}
Tomofumi Misaka, Yuko Niimura, Akiomi Yoshihisa, Kento Wada, Yusuke Kimishima,
  Tetsuro Yokokawa, Satoshi Abe, Masayoshi Oikawa, Takashi Kaneshiro, Atsushi
  Kobayashi, et~al.
\newblock Clinical impact of sleep-disordered breathing on very short-term
  blood pressure variability determined by pulse transit time.
\newblock \emph{Journal of Hypertension}, 38\penalty0 (9):\penalty0 1703--1711,
  2020.

\bibitem[Phan et~al.(2019)Phan, Andreotti, Cooray, Ch{\'e}n, and
  De~Vos]{phan2019seqsleepnet}
Huy Phan, Fernando Andreotti, Navin Cooray, Oliver~Y Ch{\'e}n, and Maarten
  De~Vos.
\newblock {SeqSleepNet: end-to-end hierarchical recurrent neural network for
  sequence-to-sequence automatic sleep staging}.
\newblock \emph{IEEE Transactions on Neural Systems and Rehabilitation
  Engineering}, 27\penalty0 (3):\penalty0 400--410, 2019.

\bibitem[Pitson et~al.(1994)Pitson, Chhina, Knijn, Van~Herwaaden, and
  Stradling]{pitson1994changes}
D~Pitson, N~Chhina, S~Knijn, M~Van~Herwaaden, and J~Stradling.
\newblock Changes in pulse transit time and pulse rate as markers of arousal
  from sleep in normal subjects.
\newblock \emph{Clinical science (London, England: 1979)}, 87\penalty0
  (2):\penalty0 269--273, 1994.

\bibitem[Pitson et~al.(1998)]{pitson1998value}
DJ~Pitson et~al.
\newblock Value of beat-to-beat blood pressure changes, detected by pulse
  transit time, in the management of the obstructive sleep apnoea/hypopnoea
  syndrome.
\newblock \emph{European Respiratory Journal}, 12\penalty0 (3):\penalty0
  685--692, 1998.

\bibitem[Punjabi(2008)]{punjabi2008epidemiology}
Naresh~M Punjabi.
\newblock {The epidemiology of adult obstructive sleep apnea}.
\newblock \emph{Proceedings of the American Thoracic Society}, 5\penalty0
  (2):\penalty0 136--143, 2008.

\bibitem[Qian et~al.(2021)Qian, Qiu, He, Lu, Lin, Xu, Zhu, Liu, Li, Cao,
  et~al.]{qian2021review}
Xiangyu Qian, Ye~Qiu, Qingzu He, Yuer Lu, Hai Lin, Fei Xu, Fangfang Zhu,
  Zhilong Liu, Xiang Li, Yuping Cao, et~al.
\newblock {A review of methods for sleep arousal detection using
  polysomnographic signals}.
\newblock \emph{Brain sciences}, 11\penalty0 (10):\penalty0 1274, 2021.

\bibitem[Quan et~al.(1997)Quan, Howard, Iber, Kiley, Nieto, O'Connor, Rapoport,
  Redline, Robbins, Samet, et~al.]{quan1997sleep}
Stuart~F Quan, Barbara~V Howard, Conrad Iber, James~P Kiley, F~Javier Nieto,
  George~T O'Connor, David~M Rapoport, Susan Redline, John Robbins, Jonathan~M
  Samet, et~al.
\newblock {The sleep heart health study: design, rationale, and methods}.
\newblock \emph{Sleep}, 20\penalty0 (12):\penalty0 1077--1085, 1997.

\bibitem[Randerath et~al.(2018)Randerath, Bassetti, Bonsignore, Farre,
  Ferini-Strambi, Grote, Hedner, Kohler, Martinez-Garcia, Mihaicuta,
  et~al.]{randerath2018challenges}
Winfried Randerath, Claudio~L Bassetti, Maria~R Bonsignore, Ramon Farre, Luigi
  Ferini-Strambi, Ludger Grote, Jan Hedner, Malcolm Kohler, Miguel-Angel
  Martinez-Garcia, Stefan Mihaicuta, et~al.
\newblock {Challenges and perspectives in obstructive sleep apnoea: report by
  an ad hoc working group of the Sleep Disordered Breathing Group of the
  European Respiratory Society and the European Sleep Research Society}.
\newblock \emph{European respiratory journal}, 52\penalty0 (3), 2018.

\bibitem[Raschke and Fischer(1997)]{raschke1997arousal}
F~Raschke and J~Fischer.
\newblock {“Arousal” in der Schlafmedizin.}
\newblock \emph{Somnologie}, 1\penalty0 (2), 1997.

\bibitem[Scharw{\"a}chter and M{\"u}ller(2020)]{scharwachter2020statistical}
Erik Scharw{\"a}chter and Emmanuel M{\"u}ller.
\newblock {Statistical evaluation of anomaly detectors for sequences}.
\newblock \emph{arXiv preprint arXiv:2008.05788}, 2020.

\bibitem[Schwartz and Moxley(2006)]{schwartz2006potential}
Daniel~J Schwartz and Pat Moxley.
\newblock {On the potential clinical relevance of the length of arousals from
  sleep in patients with obstructive sleep apnea.}
\newblock \emph{Journal of Clinical Sleep Medicine: JCSM: Official Publication
  of the American Academy of Sleep Medicine}, 2\penalty0 (2):\penalty0
  175--180, 2006.

\bibitem[Shahrbabaki et~al.(2021)Shahrbabaki, Linz, Hartmann, Redline, and
  Baumert]{shahrbabaki2021sleep}
Sobhan~Salari Shahrbabaki, Dominik Linz, Simon Hartmann, Susan Redline, and
  Mathias Baumert.
\newblock {Sleep arousal burden is associated with long-term all-cause and
  cardiovascular mortality in 8001 community-dwelling older men and women}.
\newblock \emph{European heart journal}, 42\penalty0 (21):\penalty0 2088--2099,
  2021.

\bibitem[S{\o}rb{\o} and Ruocco(2023)]{sorbo2023navigating}
Sondre S{\o}rb{\o} and Massimiliano Ruocco.
\newblock {Navigating the metric maze: A taxonomy of evaluation metrics for
  anomaly detection in time series}.
\newblock \emph{Data Mining and Knowledge Discovery}, pages 1--42, 2023.

\bibitem[Vasey et~al.(2022)Vasey, Nagendran, Campbell, Clifton, Collins,
  Denaxas, Denniston, Faes, Geerts, Ibrahim, et~al.]{vasey2022reporting}
Baptiste Vasey, Myura Nagendran, Bruce Campbell, David~A Clifton, Gary~S
  Collins, Spiros Denaxas, Alastair~K Denniston, Livia Faes, Bart Geerts,
  Mudathir Ibrahim, et~al.
\newblock {Reporting guideline for the early stage clinical evaluation of
  decision support systems driven by artificial intelligence: DECIDE-AI}.
\newblock \emph{bmj}, 377, 2022.

\bibitem[Wetter et~al.(2012)Wetter, Popp, Arzt, and
  Pollm{\"a}cher]{wetter2012elsevier}
Thomas-Christian Wetter, Roland Popp, Michael Arzt, and Thomas Pollm{\"a}cher.
\newblock \emph{{ELSEVIER ESSENTIALS Schlafmedizin: Das Wichtigste f{\"u}r
  {\"A}rzte aller Fachrichtungen}}.
\newblock Elsevier Health Sciences, 2012.

\bibitem[Wienhausen-Wilke and Kraft(2024)]{wienhausen2024computer}
Wienhausen-Wilke and Kraft.
\newblock {Computer-aided diagnostics of sleep-related arousals on the basis of
  pulse wave analyses}.
\newblock \url{https://drks.de/search/en/trial/DRKS00033641}, 2024.
\newblock [Accessed: 2024-08-15].

\bibitem[Zan and Yildiz(2023)]{zan2023multi}
Hasan Zan and Abdulnas{\i}r Yildiz.
\newblock {Multi-task learning for arousal and sleep stage detection using
  fully convolutional networks}.
\newblock \emph{Journal of Neural Engineering}, 20\penalty0 (5):\penalty0
  056034, 2023.

\bibitem[Zhang et~al.(2018)Zhang, Cui, Mueller, Tao, Kim, Rueschman, Mariani,
  Mobley, and Redline]{zhang2018national}
Guo-Qiang Zhang, Licong Cui, Remo Mueller, Shiqiang Tao, Matthew Kim, Michael
  Rueschman, Sara Mariani, Daniel Mobley, and Susan Redline.
\newblock {The National Sleep Research Resource: towards a sleep data commons}.
\newblock \emph{Journal of the American Medical Informatics Association},
  25\penalty0 (10):\penalty0 1351--1358, 2018.

\end{thebibliography}
}

\section*{Checklist}

\begin{enumerate}

\item For all authors...
\begin{enumerate}
\item Do the main claims made in the abstract and introduction accurately reflect the paper's contributions and scope?
\answerYes{See Section~\ref{sec:discussion}.}
\item Did you describe the limitations of your work?
\answerYes{See Section~\ref{sec:limitations}.
\item Did you discuss any potential negative societal impacts of your work?}
\answerNo{We do not anticipate any negative societal impacts.}
\item Have you read the ethics review guidelines and ensured that your paper conforms to them?
\answerYes{The supplementary material (Appendix~\ref{appendix:supp}) includes a detailed discussion of ethical considerations in the \textit{Datasheet} section.}
\end{enumerate}

\item If you are including theoretical results...
\begin{enumerate}
\item Did you state the full set of assumptions of all theoretical results?
\answerNA{Not applicable}
\item Did you include complete proofs of all theoretical results?
\answerNA{Not applicable}
\end{enumerate}

\item If you ran experiments (e.g. for benchmarks)...
\begin{enumerate}
\item Did you include the code, data, and instructions needed to reproduce the main experimental results (either in the supplemental material or as a URL)?
\answerNo{While the code for training and evaluation is proprietary, we included pseudo-code and detailed explanations of the ALPEC framework (Section~\ref{sec:evaluation_framework}) along with instructions on running baseline training and evaluation schemes (Sections \ref{sec:training_continuous_segmentation} and \ref{sec:baseline_approaches}). The code for these baselines is available at \url{https://github.com/rfonod/deepsleep2} (DeepSleep 2.0, MIT license) and from the work of~\citet{loning2019sktime} (sktime, BSD-3-Clause license). Details on data preprocessing and hyperparameter choices are in Appendices~\ref{appendix:data_preprocessing} and~\ref{appendix:hyperparameter_tuning}. Data from the CPS dataset is released on PhysioNet~\citep{kraft2024cps}. The PhysioNet documentation includes Croissant~\citep{akhtar2024croissant} specifications and data loading instructions, as referenced in the supplementary material (Appendix~\ref{appendix:supp}).}
\item Did you specify all the training details (e.g., data splits, hyperparameters, how they were chosen)?
\answerYes{The chosen data splits are noted with the results (Section \ref{sec:comparison_evaluation_schemes}), defined in Appendix~\ref{appendix:data_folds} and on the PhysioNet page of the CPS dataset~\citep{kraft2024cps}. Hyperparameter choices are motivated in Sections~\ref{sec:training_continuous_segmentation} and~\ref{sec:evaluation_framework}, listed in Table~\ref{tab:hyperparameters}, with ablation studies in Appendix~\ref{appendix:ablation_studies}.}
\item Did you report error bars (e.g., with respect to the random seed after running experiments multiple times)?
\answerYes{See Section~\ref{sec:comparison_evaluation_schemes}.}
\item Did you include the total amount of compute and the type of resources used (e.g., type of GPUs, internal cluster, or cloud provider)?
\answerYes{This information is included in the supplementary material (Appendix~\ref{appendix:supp}), along with an estimation of the total kgCO$_2$eq emissions.}
\end{enumerate}

\item If you are using existing assets (e.g., code, data, models) or curating/releasing new assets...
\begin{enumerate}
\item If your work uses existing assets, did you cite the creators?
\answerYes{See Section~\ref{sec:training_continuous_segmentation} for DeepSleep 2.0, Section~\ref{sec:baseline_approaches} for sktime, Appendix~\ref{appendix:physionet_challenge} for the 2018 PhysioNet Challenge Dataset and the supplementary material (Appendix~\ref{appendix:supp}) for Croissant~\citep{akhtar2024croissant} and SweetViz~\citep{bertrand2020sweetviz}.}
\item Did you mention the license of the assets?
\answerYes{We mentioned the licenses of all aforementioned assets in the above sections.}
\item Did you include any new assets either in the supplemental material or as a URL?
\answerYes{We release the Comprehensive Polysomnography (CPS) dataset~\citep{kraft2024cps} together with code and instructions on how to load the data. We reference them in multiple sections.}
\item Did you discuss whether and how consent was obtained from people whose data you're using/curating?
\answerYes{This is discussed in the \textit{Datasheet} section of the supplementary material (Appendix~\ref{appendix:supp}).}
\item Did you discuss whether the data you are using/curating contains personally identifiable information or offensive content?
\answerYes{This is discussed in the \textit{Datasheet} section of the supplementary material (Appendix~\ref{appendix:supp}).}
\end{enumerate}

\item If you used crowdsourcing or conducted research with human subjects...
\begin{enumerate}
\item Did you include the full text of instructions given to participants and screenshots, if applicable?
\answerNA{There were no specific instructions given to patients as their data was collected during routine clinical procedures, with the addition of one standardized questionnaire (Pittsburgh Sleep Quality Index) to the standard procedure.}
\item Did you describe any potential participant risks, with links to Institutional Review Board (IRB) approvals, if applicable?
\answerYes{We stated that there are no medical risks associated with participation and highlighted potential data confidentiality risks. The clinical study was approved by the ethics committee of the local ethics board.}
\item Did you include the estimated hourly wage paid to participants and the total amount spent on participant compensation?
\answerYes{Participants were not recruited as data was collected during routine clinical procedures. However, we note the compensation for a student tasked with data collection in the Datasheet section of the supplementary material (Appendix~\ref{appendix:supp}).}
\end{enumerate}

\end{enumerate}

\newpage
\begin{appendix}

\section{Additional experiments on further training and evaluation schemes}\label{appendix:physionet_challenge}
In this section, we explore the transition from full arousal event detection (FED) to arousal onset detection using continuous segmentation (CS) training approaches, evaluated under both the traditional pointwise scheme and our ALPEC framework. For an overview of all training and evaluation schemes, refer to Appendix~\ref{appendix:schemes_comparison}.

The 2018 PhysioNet Challenge Dataset, licensed under the Open Data Commons Attribution License v1.0, is a notable publicly available resource that includes polysomnographic (PSG) data from 1,983 patients at Massachusetts General Hospital's Sleep Lab, with labels provided for 994 subjects~\citep{ghassemi2018you, goldberger2000physiobank}. It adheres to AASM guidelines and includes 13 data channels (six EEG, EOG, EMG at the chin, respiratory at chest and abdomen, ECG, SaO\textsubscript{2}, and airflow) annotated with various sleep stages and arousal categories.

The 2018 PhysioNet Challenge dataset provides a basis for training and comparing a full event detection (FED) baseline due to its arousal annotations with both meaningful start and end points. We randomly partitioned the 994 samples into a training set of 795 samples and a test set of 199 samples. Utilizing all 13 channels, training proceeded until either early stopping criteria were met or 50 epochs were completed. The results are shown in Table \ref{tab:results_physionet}.

\begin{table}[!htbp]
\caption{\textbf{Arousal Detection Performance on the 2018 PhysioNet Challenge Dataset}~\citep{ghassemi2018you}. Performance metrics are presented for the original challenge task targeting only Respiratory Effort-Related Arousals (\textit{RERA}) and for the most frequently annotated arousal types (\textit{Most freq.}), which include RERA, Hypopnea, Central apnea, and Obstructive apnea. The training approaches (\textit{FED}: Full Event Detection, \textit{POD}: Point-based Onset Detection, \textit{IOD}: Interval-based Onset Detection) employ the DeepSleep method discussed in Section~\ref{sec:training_continuous_segmentation}. Evaluation methods (\textit{PE}: Pointwise Evaluation, \textit{ALPEC}: Approximate Localization and Precise Event Count) are further explained in Sections~\ref{sec:evaluation_framework} and~\ref{sec:baseline_approaches}. Performance metrics are reported as mean values over all test subjects, employing cross-subject validation. All models undergo five training iterations, with results averaged and presented alongside 95\% confidence intervals in brackets, assuming t-distributed mean values. Key findings highlight that POD is ineffective with the DeepSleep approach. In contrast, ALPEC reveals that DeepSleep for IOD performs comparably to the clinically more demanding FED baseline, unlike PE.}\label{tab:results_physionet}
\begin{center}
\setlength{\tabcolsep}{4pt} %
\begin{tabular}{llllllll}
\toprule
\multirow{2}{*}{Targets} & \multirow{2}{*}{Training approach} & \multicolumn{3}{c}{PE (baseline)} & \multicolumn{3}{c}{ALPEC (our)}\\
\cmidrule(lr){3-5}
\cmidrule(lr){6-8}
& & $\bar{\text{Precision}}$ & $\bar{\text{Recall}}$ & $\bar{\text{F2}}$ & $\bar{\text{Precision}}$ & $\bar{\text{Recall}}$ & $\bar{\text{F2}}$ \\
\midrule
\multirow{3}{*}{RERA}
& IOD (our) & \errfmt{0.13}{4} & \errfmt{0.47}{7} & \errfmt{0.30}{3} & \errfmt{0.20}{6} & \errfmt{0.63}{12} & \errfmt{0.42}{4} \\

& POD (naive baseline) & 7e-6 & 0.94 & 3.5e-5 & 0.00 & 0.00 & 0.00 \\
& FED (baseline) & \errfmt{0.17}{5} & \errfmt{0.49}{17} & \errfmt{0.35}{8} & \errfmt{0.23}{15} & \errfmt{0.59}{20} & \errfmt{0.41}{6} \\
\midrule
\multirow{3}{*}{\makecell[tl]{Most\\freq.}}
& IOD (our) & \errfmt{0.33}{6} & \errfmt{0.71}{10} & \errfmt{0.57}{3} & \errfmt{0.53}{6} & \errfmt{0.85}{6} & \errfmt{0.76}{3} \\
& POD (naive baseline) & 3.4e-5 & 1.00 & 	
1.67e-4 & 0.00 & 0.00 & 0.00\\
& FED (baseline) & \errfmt{0.54}{4} & \errfmt{0.67}{8} & \errfmt{0.64}{5} & \errfmt{0.60}{4} & \errfmt{0.77}{5} & \errfmt{0.73}{2} \\
\bottomrule
\end{tabular}
\end{center}
\end{table}

Our results indicate that point-based onset detection (POD) using the DeepSleep approach for continuous segmentation is infeasible. Due to the sparsity of labels and noise in the onset annotations, the model is unable to relate meaningful patterns to arousal onsets. In pointwise evaluation (PE), a very low decision threshold results in high recall but low precision, leading to the best possible F2 score, which remains close to zero. ALPEC offers a more realistic assessment of the model's performance by discarding excessively long predicted intervals.
However, if a model were to make point-predictions for arousal onsets within the temporal tolerance buffer of ground-truth onset points, ALPEC is expected to perform adequately, whereas pointwise evaluation would penalize predictions that are off by even one point. Thus, ALPEC enables the utilization and appropriate evaluation of other point-based detection approaches that have not been used before, such as methods for changepoint detection in time series~\citep{aminikhanghahi2017survey}.

Moreover, interval-based onset detection (IOD) performs comparably to the FED baseline when measured by ALPEC, demonstrating that detecting arousal onsets rather than full events can be equally effective. Conversely, pointwise evaluation asserts a substantial performance discrepancy, underscoring the importance of choosing an appropriate evaluation framework.

As a final remark, the rather large confidence intervals in the precision and recall results stem from the variance in selected decision thresholds for individual runs, as shown in Table~\ref{tab:study_of_thresholds}.

\begin{table}[!htbp]
\caption{\textbf{Comparison of selected decision thresholds and their effects}. This table presents the decision thresholds for five individual runs with different random seeds, leading to the results for the \textit{RERA} targets using \textit{FED} training and \textit{ALPEC} evaluation shown in Table~\ref{tab:results_physionet}.}\label{tab:study_of_thresholds}
\begin{center}
\begin{tabular}{llllll}
\toprule
Run & Decision Threshold & $\bar{\text{Precision}}$ & $\bar{\text{Recall}}$ & $\bar{\text{F1}}$ & $\bar{\text{F2}}$ \\
\midrule
1 & 0.22 & 0.44 & 0.42 & 0.43 & 0.43 \\
2 & 0.03 & 0.17 & 0.82 & 0.47 & 0.47 \\
3 & 0.07 & 0.18 & 0.68 & 0.44 & 0.44 \\
4 & 0.13 & 0.19 & 0.48 & 0.37 & 0.37 \\
5 & 0.12 & 0.15 & 0.53 & 0.35 & 0.35 \\
\bottomrule
\end{tabular}
\end{center}
\end{table}

These thresholds are based on samples from the training fold and are automatically selected to maximize the F2 score. Comparing runs 1 and 2, for example, shows that a lower decision threshold results in higher recall but lower precision, as expected, while the resulting F2 scores are comparable.

\section{Ablation studies}\label{appendix:ablation_studies}
In this section, we present ablation studies conducted on the CPS dataset to evaluate the impact of various hyperparameters on the performance of the DeepSleep method for arousal onset detection and the ALPEC framework for performance evaluation. The hyperparameters considered are the smoothing window size ($w$), the interval length for interval-based onset detection (IOD) ($l$), the maximum interval duration ($d$), the minimum interval distance ($\delta$), and the buffer size for the ALPEC evaluation framework. For a description of the parameters, see Table~\ref{tab:table_of_notation}. We use the DeepSleep model with a univariate \textit{C3:A2} channel. All models are trained using the same training and test split as in Section~\ref{sec:main_results}.

\begin{table}[!htb]
\caption{\textbf{Ablation studies over hyperparameters} used in interval-based onset detection (IOD) training (a, b) and ALPEC (c, d, e). For a description of the parameters, see Table~\ref{tab:table_of_notation}. D1 from Table~\ref{tab:comparion_evaluation_schemes} is the base model used. Parameter choices for all runs, if not tuned, are marked in bold, which are the same as for D1 in the main part of this paper.}
\label{tab:ablation_studies}
\begin{center}

\setlength{\tabcolsep}{3pt} %
\begin{adjustbox}{valign=t,minipage=.33\linewidth}
\captionof*{table}{(a) $w$: Smoothing window}
\vspace{1em}
\begin{tabular}{llll}
\toprule
$w$ & $\bar{\text{Precision}}$ & $\bar{\text{Recall}}$ & $\bar{\text{F2}}$ \\
\midrule
none & 0.47 & 0.67 & 0.60\\
1 & 0.59 & 0.71 & 0.66\\
2 & 0.53 & 0.73 & 0.65\\
\textbf{3} & 0.47 & 0.80 & 0.68\\
4 & 0.45 & 0.86 & 0.70\\
5 & 0.44 & 0.82 & 0.67\\
\bottomrule
\end{tabular}
\end{adjustbox}%
\begin{adjustbox}{valign=t,minipage=.33\linewidth}
\captionof*{table}{(b) $l$: Interval length for IOD}
\vspace{1em}
\centering
\begin{tabular}{llll}
\toprule
$l$ & $\bar{\text{Precision}}$ & $\bar{\text{Recall}}$ & $\bar{\text{F2}}$ \\
\midrule
2 & 0.39 & 0.78 & 0.62\\
6 & 0.49 & 0.67 & 0.61\\
\textbf{10} & 0.45 & 0.80 & 0.68\\
14 & 0.47 & 0.80 & 0.68\\
20 & 0.42 & 0.81 & 0.66\\
30 & 0.47 & 0.59 & 0.55\\
60 & 0.36 & 0.50 & 0.45\\
\bottomrule
\end{tabular}
\end{adjustbox}%
\begin{adjustbox}{valign=t,minipage=.33\linewidth}
\captionof*{table}{(c) $d$: Max. interval duration}
\vspace{1em}
\centering
\begin{tabular}{llll}
\toprule
$d$ & $\bar{\text{Precision}}$ & $\bar{\text{Recall}}$ & $\bar{\text{F2}}$ \\
\midrule
10 & 0.57 & 0.19 & 0.22\\
30 & 0.50 & 0.74 & 0.66\\
\textbf{60} & 0.47 & 0.80 & 0.68\\
90 & 0.43 & 0.77 & 0.65\\
120 & 0.40 & 0.86 & 0.67\\
none & 0.45 & 0.82 & 0.68\\
\bottomrule
\end{tabular}
\end{adjustbox}%
\vspace{0.5em}
\begin{adjustbox}{valign=t,minipage=.49\linewidth}
\captionof*{table}{(d) $\delta$: Min. interval distance}
\vspace{1em}
\centering
\begin{tabular}{llll}
\toprule
$\delta$ & $\bar{\text{Precision}}$ & $\bar{\text{Recall}}$ & $\bar{\text{F2}}$ \\
\midrule
0 & 0.48 & 0.76 & 0.66\\
5 & 0.50 & 0.74 & 0.66\\
\textbf{10} & 0.47 & 0.80 & 0.68\\
15 & 0.37 & 0.86 & 0.66\\
20 & 0.37 & 0.58 & 0.50\\
\bottomrule
\end{tabular}
\end{adjustbox}
\begin{adjustbox}{valign=t,minipage=.49\linewidth}
\captionof*{table}{(e) $b$: Buffer size with $b=b^\text{before}=b^\text{after}$}
\vspace{1em}
\centering
\begin{tabular}{llll}
\toprule
$b$ & $\bar{\text{Precision}}$ & $\bar{\text{Recall}}$ & $\bar{\text{F2}}$ \\
\midrule
0 & 0.51 & 0.69 & 0.63\\
5 & 0.47 & 0.70 & 0.61\\
10 & 0.46 & 0.75 & 0.65\\
\textbf{15} & 0.47 & 0.80 & 0.68\\
20 & 0.43 & 0.82 & 0.68\\
25 & 0.59 & 0.66 & 0.63\\
\bottomrule
\end{tabular}
\end{adjustbox}
\end{center}
\end{table}

We can see from Table~\ref{tab:ablation_studies} that most parameters have a moderate effect on performance metrics within the ranges tested. The most significant drop in performance occurs with low values of the maximum allowed interval distance $d$, which is expected since removing many events leads to a high number of false negatives. Values higher than $d=60s$ make no significant difference to $d=60s$, indicating that our ML approach produces reasonably short predicted intervals. We also see that smoothing (a), merging of intervals (d), and utilizing a buffer (e) all lead to performance improvements. Smoothing and merging actually affect the predicted intervals, whereas the buffer only affects the evaluation by relaxing the locality requirement.

\section{Data preprocessing}\label{appendix:data_preprocessing}

For our experiments, all raw data channels undergo third-order Butterworth bandpass filtering to remove noise. Critical frequencies for the Butterworth bandpass filter for the different data modalities are listed in Table \ref{tab:bandpass_filter}.

\begin{table}[!htb]
\caption{Critical frequencies for the bandpass filter for different modalities}\label{tab:bandpass_filter}
\begin{center}
\begin{tabular}{llll}
\toprule
\textbf{Modality} & \textbf{Channels} & \textbf{Lower freq. [Hz]} & \textbf{Upper freq. [Hz]} \\
\cmidrule(lr){1-4}
EEG & \makecell[tl]{C4:A1, C3:A2, F4:A1, O2:A1,\\A1, A2, C3, C4, F4, O2} & 0.2 & 35 \\
EOG & \makecell[tl]{EOGl, EOGL:A1, EOGL:A2,\\EOGr, EOGr:A1, EOGr:A2} & 0.2 & 35 \\
EMG & EMG+, EMG-, EMG & 10 & 127 \\
ECG & ECG 2 & 0.2 & 127 \\
Respiratory & Pressure Flow, Thermal Flow & 0.001 & 15 \\
Snore & Snoring Pressure, Snoring Sound & 20 & 127 \\
PPG & Pleth & 0.5 & 5 \\
\bottomrule
\end{tabular}
\end{center}
\end{table}

The raw data channels are then normalized using z-score normalization. Derived channels are upsampled to 256 Hz using repeated values and scaled to a range of $[0, 1]$ via min-max normalization. Channels are padded symmetrically to a fixed length of $n=2^{23}$, or approximately 9 hours, to accommodate the longest recording. Magnitude scaling is applied randomly between 0.8 and 1.25 during training to enhance model generalization~\cite{li2021deepsleep}.

All nominal event data are encoded as binary features, with each event type represented as a separate feature. For Sleep Profile and Body Position events, we utilize a one-hot encoding scheme to represent the different classes.

\section{Hyperparameter tuning details}
\label{appendix:hyperparameter_tuning}
In this section, we provide an overview of the selected hyperparameters and perform preliminary experiments with the DeepSleep approach for continuous segmentation on unimodal data channels to find a good set of input channels for final model candidates. All models are trained on the training set and evaluated on a fixed validation set. See Appendix \ref{appendix:data_folds} for details on the data splits. Table~\ref{tab:hyperparameters} contains an overview of the hyperparameters used in this work.

\begin{table}[!htbp]
\caption{\textbf{Choices for Hyperparameters}. Values in seconds are multiplied by the fixed sampling rate of 256 Hz. For further explanations of the meaning of the symbols, see Table~\ref{tab:table_of_notation}}
\label{tab:hyperparameters}
\begin{center}
\begin{tabular}{llll}
\toprule
Context & Parameter & Value & Explanation\\
\cmidrule(lr){1-4}

\multirow{1}{*}{Data (CPS)}
& $|T|$ & $64$ & Number of subjects in the training set\\
& $|V|$ & $28$ & Number of subjects in the validation set\\
& $|E|$ & $14$ & Number of subjects in the test set\\

\cmidrule(lr){1-4}

\multirow{1}{*}{Training}
& $n$ & $2^{23}$ & Number of padded data points per channel\\
& $s$ & $30s$ & Window size for window-based classification\\
& $\omega$ & 3s & Smoothing window for continuous segmentation\\
& $l$ & 10s & Interval length for IOD\\
& epochs & 100 & Maximum number of training epochs\\
& batch size & 1 & Number of subjects per batch\\
\cmidrule(lr){1-4}

\multirow{1}{*}{ALPEC}
& $d$ & 60s & Maximum interval duration\\
& $\delta$ & 10s & Minimum interval distance\\
& $b^\text{before}$ & 15s & Left buffer size\\
& $b^\text{after}$ & 15s & Right buffer size\\
\bottomrule

\end{tabular}
\end{center}
\end{table}

\paragraph{Selection of input channels}

We perform two baseline sets of tuning runs: one on raw channels (see Table \ref{tab:psg_raw_channels}) and another on derived channels and a promising selection of event channels (see Tables \ref{tab:psg_derived_channels} and \ref{tab:annotated_events}). From the raw channels, we leave out the \textit{Battery} and \textit{REM Confidence} channels since we expect those to be irrelevant for arousal detection. Also, we only use one channel each from the EEG, EMG, and EOG groups. We split the categorical event channels \textit{Sleep Profile} and \textit{Body Position} into singular channels using a one-hot encoded representation of the categories. For simplicity, we will refer to both the derived and event channels as \textit{derived} channels from here on.
All results from these runs are shown in Table \ref{tab:tuning_unimodal_channels}.

\begin{table}[!htbp]
\caption{\textbf{Model tuning with unimodal channel data} for arousal detection on the CPS dataset using the DeepSleep model for continuous segmentation. The raw channels are explained in Table \ref{tab:psg_raw_channels}, derived channels in Tables \ref{tab:psg_derived_channels} and \ref{tab:annotated_events}. \textit{D3} and \textit{D4} are selections of channels that are used for model candidates in the main part of this paper.}\label{tab:tuning_unimodal_channels}
\begin{center}

\begin{adjustbox}{valign=t,minipage=.49\linewidth}
\captionof*{table}{(a) Unimodal training on raw channels}
\vspace{1em}
\begin{tabular}{llll}
\toprule
Channel & $\bar{F}_2$ & D3 & D4\\
\midrule

C3:A2 & 0.60 & & \checkmark\\
Pressure Flow & 0.56 & \checkmark & \checkmark\\
RIP.Abdom & 0.53 & \checkmark & \checkmark\\
EMG & 0.53 & & \checkmark\\
Sum RIPs & 0.52 & \checkmark & \checkmark\\
EOGl & 0.50 & & \checkmark\\
Pulse & 0.48 & \checkmark & \checkmark\\
RIP.Thrx & 0.46 & \checkmark & \checkmark\\
Pleth & 0.45 & \checkmark & \checkmark\\
Snoring Pressure & 0.45 & \checkmark & \checkmark\\
Thermal Flow & 0.43 & \checkmark & \checkmark\\
PLMl & 0.37 & \checkmark & \checkmark\\
ECG 2 & 0.36 & \checkmark & \checkmark\\
Light & 0.34 & \checkmark & \checkmark\\
SPO2 & 0.33 & \checkmark & \checkmark\\
Snoring Sound & 0.30 & \checkmark & \checkmark\\
Motion & 0.28 & \checkmark & \checkmark\\

\bottomrule
\end{tabular}

\end{adjustbox}
\begin{adjustbox}{valign=t,minipage=.49\linewidth}
\captionof*{table}{(b) Unimodal training on derived channels}
\vspace{1em}
\begin{tabular}{llll}
\toprule
Channel & $\bar{F}_2$ & D3 & D4\\
\midrule

Average Frequency Value & 0.55 & & \checkmark\\
Hypopnea & 0.53 & \checkmark & \checkmark\\
Sigma FFT & 0.50 & & \checkmark\\
Heart Rate & 0.48 & & \\
RR Interval & 0.45 & \checkmark & \checkmark\\
Delta FFT & 0.45 & & \checkmark\\
Alpha+Beta FFT & 0.44 & & \checkmark\\
PTT Raw & 0.44 & \checkmark & \checkmark\\
HRV LF & 0.43 & \checkmark & \checkmark\\
Diastol & 0.43 & \checkmark & \checkmark\\
Obstruction & 0.43 & \checkmark & \checkmark\\
Systol PTT & 0.42 & \checkmark & \checkmark\\
Sleep Profile & 0.42 & & \checkmark\\
Syst & 0.42 & \checkmark & \checkmark\\
Diastol PTT & 0.41 & \checkmark & \checkmark\\
RR & 0.41 & \checkmark & \checkmark\\
SVB & 0.41 & \checkmark & \checkmark\\
Phase Angle & 0.41 & \checkmark & \checkmark\\
Light & 0.33 & & \\
SpO2 & 0.31 & & \\
Activity & 0.28 & \checkmark & \checkmark\\
Integral EMG & 0.26 & & \checkmark\\
HRV HF & 0.25 & \checkmark & \checkmark\\
Obstructive Apnea & 0.17 & \checkmark & \checkmark\\
Apnea & 0.09 & \checkmark & \checkmark\\
Body Position & 0.04 & & \\
Central Apnea & 0.04 & & \\

\bottomrule
\end{tabular}
\end{adjustbox}
\end{center}
\end{table}

Tuning all possible combinations of raw and derived channels from Tables \ref{tab:psg_raw_channels}, \ref{tab:psg_derived_channels}, and \ref{tab:annotated_events} would be computationally very demanding, even when restricting ourselves to the most discriminative channels. From explorative experiments, we learned that the performance of DeepSleep generally increases when using additional channels. Therefore, we selected four combinations of channels as model candidates for the final evaluation in the main section of this work, denoted with a \textit{D} for \textit{DeepSleep}:

\begin{enumerate}
\item D1, using only the channel \textit{C3:A2}, which yielded the best performance in Table \ref{tab:tuning_unimodal_channels} and is the required choice for manual arousal detection according to the AASM guidelines~\citep{berry2012rules}.
\item D2, using modalities often selected for arousal detection in related work (see Table~\ref{tab:related_work}), namely \textit{C3:A2}, \textit{EOGl}, and \textit{EMG}.
\item D3, using a selection of channels that do not rely on EEG, EMG, and EOG modalities as indicated in Table~\ref{tab:tuning_unimodal_channels} in the \textit{D3} column.
\item D4, using all channels from Table~\ref{tab:tuning_unimodal_channels} except for the most underperforming channels, \textit{Body Position} and \textit{Central Apnea}, also indicated in Table~\ref{tab:tuning_unimodal_channels} in the \textit{D4} column.
\end{enumerate}

Additionally, we did not select the \textit{Heart Rate}, \textit{Light}, and \textit{SpO2} derived channels for model candidates \textit{D3} and \textit{D4} since these are very similar to the \textit{Pulse}, \textit{Light}, and \textit{SPO2} raw channels, respectively, as indicated by the similar performances in Table~\ref{tab:tuning_unimodal_channels}.

The effects of additional hyperparameters are detailed in Appendix \ref{appendix:ablation_studies}, where calculations are performed using the \textit{D1} unimodal channel selection.

\section{Schematic comparison of training and evaluation schemes}\label{appendix:schemes_comparison}
\begin{figure}[!htbp]
\begin{center}
\includegraphics[width=1.0\linewidth]{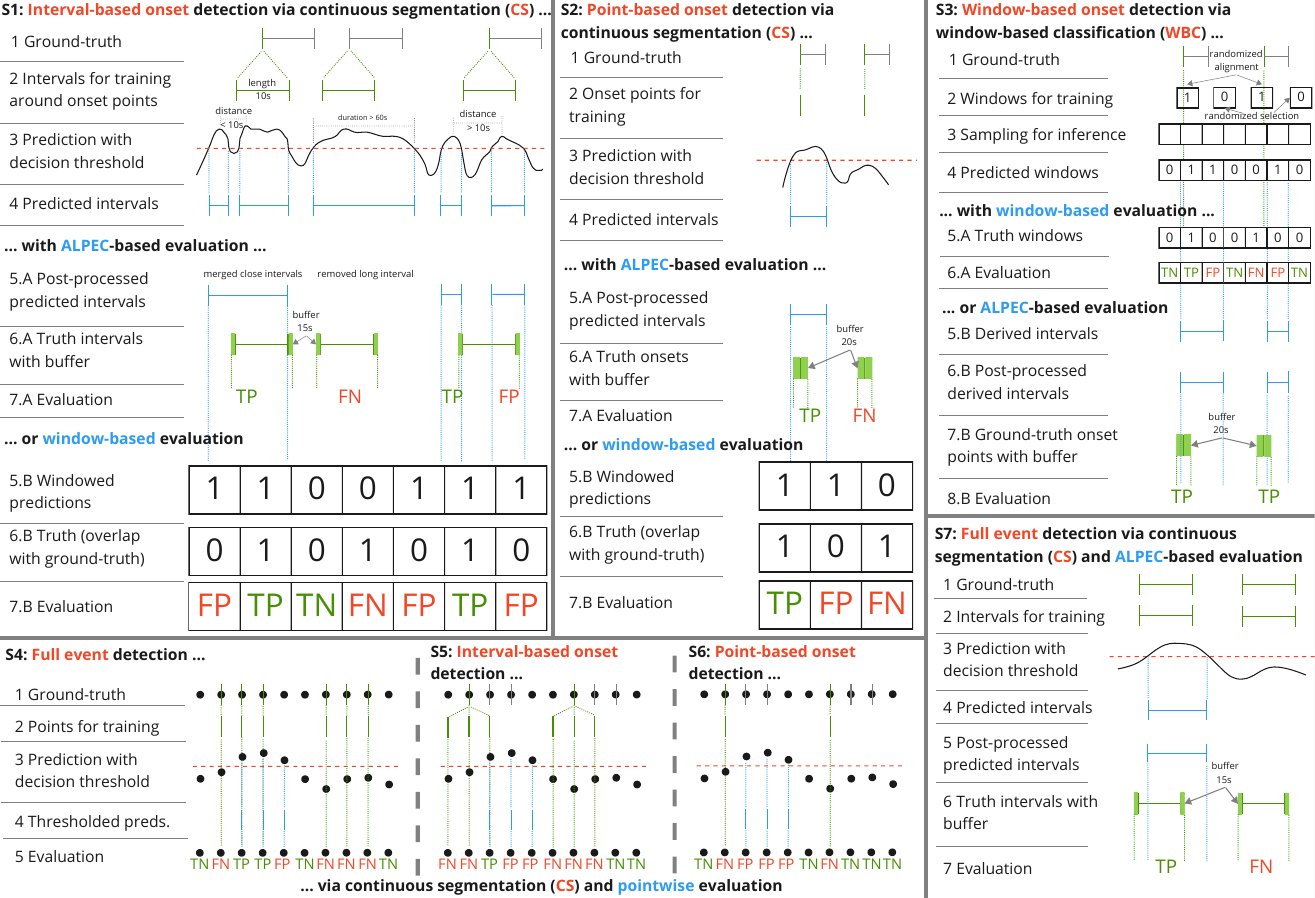}
\\
\end{center}
\caption{\textbf{Schematic illustration of different approaches for training and evaluating arousal detection models.} In schemas S1, S2, S3, and S7, lines and areas in green color represent target points of the positive class (arousal) while empty areas in between contain points of the negative class (no arousal). Lines in blue color represent points that are predicted to be in the positive class. For schemas containing pointwise evaluations (S4-S6), all points which are marked in green or blue are considered to be in the positive class while all other points are considered to be in the negative class. Names of training schemes are highlighted in red, evaluation schemes in blue. 
All sizes and dimensions are for illustrative purposes and not representative. Especially schemas containing pointwise evaluations will contain many more data points inside events/intervals (S4-S5) and between events (S4-S6). For schemas containing window-based approaches (S1-S3), each box represents a window of fixed length containing many data points, where the class identification or evaluation outcome of each point is given by the label on the box.}\label{fig:schemes}
\end{figure}

In this work, we utilize a multitude of training and evaluation approaches which are schematically illustrated in Figure~\ref{fig:schemes}. For an explanation of the schemes, we refer to Section~\ref{sec:methods} and Appendix~\ref{appendix:alpec}.

We now want to perform a more detailed conceptual comparison of our proposed Approximate Localization and Precise Event Count (ALPEC) framework with the baselines of pointwise evaluation and window-based evaluation.
We start by remembering that, from a clinical productional point-of-view, the most important aspect of arousal diagnostics is to detect the correct number of arousals and to locate them approximately correctly to enable human validation (cf. Section~\ref{sec:evaluation_framework}). Looking at the \textit{Evaluation} steps of S4-S6 in Figure~\ref{fig:schemes}, we see that pointwise evaluation is inadequate based on our requirements since it sanctions every wrong prediction point. This leads to a bias towards favoring models that strictly predict the exact labeled points which might also result in overfitting and a lack of the generalization capabilities of models that are optimized under pointwise evaluation.
When comparing the exact situation with ALPEC instead of pointwise evaluation (S7), we see that ALPEC is not concerned with single points but only close-by intervals and counts exactly one TP and one FN as would be expected in this situation from a clinical point-of-view. As \citet{sorbo2023navigating} have noted generally about pointwise evaluation, a major shortcoming is the lack of tolerance which renders it inappropriate for the evaluation of arousal detection models.

Moving on to window-based evaluation, we find a similar situation as with pointwise evaluation although less severe. Looking at S3, we see that window-based evaluation sanctions the third predicted window although adjacent to a correct prediction (the second) and sanctions the second arousal twice (one FN, one FP), for the close miss of the onset point at the border to the window with a predicted 1. To be fair, window-based onset detection could be equipped with similar domain-specific adaptations like the interval-based ALPEC. Our fundamental critique, however, is that with a windowing approach there are always technical constraints due to the window size which lead to deviations from the intended goal which can be utterly avoided by using the interval-based approach of ALPEC. It entails intrinsic flexibility, allowing it to be closely adapted to the clinical needs. Apart from this, as we have seen in Section~\ref{sec:methods_arousal_detection}, window-based classification approaches often contain many hyperparameters related to window size, overlap, and voting strategies which often extend to window-based evaluation. ALPEC contains hyperparameters of its own which, however, are less technical and instead are introduced to foster an adaptation to the productive clinical requirements.

\section{Formal description of ALPEC}\label{appendix:alpec}

We now formally introduce the procedure of our post-processing and performance evaluation framework ALPEC. A table of notation can be found in Table~\ref{tab:table_of_notation}.

\begin{table}[!htbp]
\caption{Table of notation}
\label{tab:table_of_notation}
\setlength{\tabcolsep}{4pt}
\begin{center}
\begin{tabular}{ll}

\toprule
Symbol & Meaning \\
\cmidrule(lr){1-2}
$\mathbf{x}$ & Multivariate input sequence \\
$n$ & \makecell[tl]{Number of data points contained within each input channel after padding with zeros,\\fixed to $2^{23}$}\\
$D$ & Dataset containing all subjects\\
$T$ & Training set containing a subset of subjects\\
$V$ & Validation set containing a subset of subjects\\
$p_i(\mathbf{x})$ & Probability score of the $i$-th time step in the input sequence being of the positive class\\
$p_\eta(\mathbf{x})$ & Probability score for the $\eta$-th window in the input sequence\\
$c_i(\mathbf{x})$ & Binary class prediction for the $i$-th time step\\
$N$ & Number of windows when splitting the input sequence into windows of length $s$\\
$s$ & Length of each window when splitting the input sequence into $N$ windows\\
$w$ & Window size for smoothing the probability scores\\
$f$ & Sampling frequency of the data\\
$t_k$ & \makecell[tl]{Threshold for converting probability scores to binary class predictions.\\We use 101 thresholds from 0 to 1 in steps of 0.01, i.e. $k=0,1,...,100$}\\
$\delta$ & Minimum distance in seconds for merging two adjacent predicted intervals in ALPEC\\
$C$ & Binary class predictions for the whole input sequence, i.e. $C=\{c_1, c_2, ..., c_n\}$\\
$I$ & Predicted interval in binary class predictions $C$ with start and end indices $I^\text{start}$ and $I^\text{end}$\\
$G$ & Ground-truth interval with start and end indices $G^\text{start}$ and $G^\text{end}$\\
$d$ & Maximum duration of a predicted interval before its removal in ALPEC\\
$P$ & Predicted interval with start and end indices $P^\text{start}$ and $P^\text{end}$\\
$b^\text{before}$ & Buffer before the ground-truth interval in ALPEC\\
$b^\text{after}$ & Buffer after the ground-truth interval in ALPEC\\
$G^\text{ext}$ & Extended ground-truth interval with start and end indices $G^\text{start,ext}$ and $G^\text{end,ext}$\\
$l$ & Length of the interval around an onset point for interval-based onset detection (IOD)\\

\bottomrule
\end{tabular}
\end{center}
\end{table}

If we are using a continuous segmentation approach (see S1, S2, and S7 in Figure~\ref{fig:schemes}), we start with a probability score $p_{i,\nu}(\mathbf{x})$ for each data point $i=1,...,n$ from the measurement data $\mathbf{x}$ and each subject $\nu$ with $\nu=1,...,|D|$ in the dataset $D$, where $n=2^{23}$ is the padded fixed number of data points for each input channel.
Alternatively, if we use a window-based classification (WBC) approach (see S3 in Figure~\ref{fig:schemes}), we start with probability scores $p_{\eta,\nu}(\mathbf{x})$ or binary class predictions $c_{\eta,\nu}(\mathbf{x})$, where $\eta=1,...,N$, and the data is divided into $N$ windows of equal length $s$.

When starting with probability scores, i.e., $p_{i,\nu}(\mathbf{x})$ or $p_{\eta,\nu}(\mathbf{x})$, 
we apply a threshold $t_k$ to the scores to obtain binary class predictions $c_{i,\nu,k}(\mathbf{x})$ or $c_{\eta,\nu,k}(\mathbf{x})$ respectively, where thresholds $t_k$ are selected from 0 to 1 in steps of 0.01, i.e., $k=0,1,...,100$:

\begin{equation}
c_{i,\nu,k}(\mathbf{x}) = \begin{cases}
        1 & \text{if } p_{i,\nu}(\mathbf{x}) \geq t_k\\
0 & \text{otherwise}
\end{cases}
\end{equation}
or
\begin{equation}
c_{\eta,\nu,k}(\mathbf{x}) = \begin{cases}
        1 & \text{if } p_{\eta,\nu}(\mathbf{x}) \geq t_k\\
0 & \text{otherwise}
\end{cases}
\end{equation}

In the case of WBC, we convert the window-based predictions to pointwise predictions $c_{i,\nu}(\mathbf{x})$ by assigning the prediction of the window to all data points within the window:

\begin{equation}
    c_{i,\nu}(\mathbf{x}) = c_{\eta,\nu}(\mathbf{x}) \text{ for } i \in [(\eta-1) \cdot s + 1, \eta \cdot s]
\end{equation}

At this point, both starting points (continuous segmentation and window-based classification) are synchronous again.
Next, we merge predictions less than $\delta$ seconds apart.
For ease of notation, we temporarily drop the indices $k$ and $\nu$ and parameter $\mathbf{x}$.
At first, for the sequence of binary target values $C=(c_1, c_2, ..., c_n)$, we identify the start and end indices of each predicted interval $P$ in $C$ as $P^\text{start}$ and $P^\text{end}$, respectively, where an interval starts at index $i$ if $c_i=1$ and $c_{i-1}=0$ and ends at index $j$ if $c_j=1$ and $c_{j+1}=0$ if all intermediate predictions $c_{i+1}, ..., c_{j-1}$ are equal to 1. For arousal onset detection, the distance is calculated based on the maxima of the scores of two consecutive predicted intervals. For each identified interval $P$, we find the index $m$ within the interval that maximizes the score $p_m$:

\begin{equation}
    m = \text{arg\,max}_{m \in [P^\text{start}, P^\text{end}]} s_m
\end{equation}

Two intervals $P_1$ and $P_2$, with maximum score indices $m_1$ and $m_2$, are merged if $|m_1-m_2| < \delta \cdot f$. In the case of full event detection, we merge two intervals based on their start and end points, i.e., if $|P_1^\text{start}-P_2^\text{end}| < \delta \cdot f$.
The merged interval $P^\text{merged}$ then extends from $P_1^\text{start}$ to $P_2^\text{end}$. For the merged sequence $C^\text{merged}$ we set:

\begin{equation}
    c_i^\text{merged} = \begin{cases}
        1 & \text{if } i \in \text{any} P^\text{merged}\\
        0 & \text{otherwise}
    \end{cases}
\end{equation}

The final step is to compare the predicted intervals $P$ with the ground-truth intervals $G$ with start end points $G^\text{start}$ and $G^\text{end}$In the case of point-based onset detection, $G^\text{start}=G^\text{end}$, meaning the ground-truth intervals are points in time. Our ALPEC framework, however, remains generic and can be used for any length of ground-truth intervals, thus supporting both interval-based onset detection and full event detection.

Next, we once again tailor the evaluation method to the specificities of the task by introducing two factors:
First, we define a maximum duration $d$ for the predicted intervals, which we will utilize shortly.
Second, we extend all ground-truth intervals with a buffer $b^\text{before}$ on the left $b^\text{after}$ on the right side of the interval:

\begin{equation}
    G^\text{start,ext} = \max(0, G^\text{start} - b^\text{before} \cdot f)
\end{equation}

\begin{equation}
    G^\text{end,ext} = \min(n, G^\text{end} + b^\text{after} \cdot f)
\end{equation}

Now, we can proceed with the comparison of predicted and ground-truth intervals to calculate the number of true positives (TP), false positives (FP), and false negatives (FN).
A TP is counted when a predicted interval overlaps with a ground-truth interval extended by the buffer, so that $P^\text{start} \leq G^\text{end,ext}$ and $P^\text{end} \geq G^\text{start,ext}$ and the duration of the predicted interval $P^\text{end} - P^\text{start} \leq d \cdot f$. 
A FP is counted when a predicted interval does not overlap with any extended ground-truth interval. A FN is counted when an extended ground-truth interval does not overlap with any predicted interval.
If multiple predicted intervals overlap with a single ground-truth interval, we count only one TP for the first predicted interval that overlaps with the ground-truth interval. Each additional predicted interval that overlaps with the same ground-truth interval is counted as a FP.
Conversely, when a single predicted interval overlaps with multiple ground-truth intervals, we adhere to a conservative matching strategy. A TP is counted for the first ground-truth interval matched by the predicted interval. However, if additional ground-truth intervals are overlapped by the same predicted interval but are not matched by another distinct predicted interval, those are counted as false negatives (FN). This rule is designed to penalize the prediction of overly long intervals that may inaccurately suggest a single event where multiple distinct events are present.

Now, we derive the following performance metrics for each subject $\nu$ and each threshold $t_k$, where we re-introduce the indices:

\begin{equation}
    \text{Precision}_{\nu,k} = \frac{\text{TP}_{\nu,k}}{\text{TP}_{\nu,k} + \text{FP}_{\nu,k}}
\end{equation}

\begin{equation}
    \text{Recall}_{\nu,k} = \frac{\text{TP}_{\nu,k}}{\text{TP}_{\nu,k} + \text{FN}_{\nu,k}}
\end{equation}

\begin{equation}
    \text{F1}_{\nu,k} = \frac{2 \cdot \text{Precision}_{\nu,k} \cdot \text{Recall}_{\nu,k}}{\text{Precision}_{\nu,k} + \text{Recall}_{\nu,k}}
\end{equation}

\begin{equation}
    \text{F2}_{\nu,k} = \frac{5 \cdot \text{Precision}_{\nu,k} \cdot \text{Recall}_{\nu,k}}{4 \cdot \text{Precision}_{\nu,k} + \text{Recall}_{\nu,k}}
\end{equation}

After having laid out the process to calculate the performance metrics for each subject and each threshold, we now describe how we aggregate these metrics across subjects and how we deal with data splits and thresholding.

The first step is to calculate the micro average $F2_k^\text{train}$ score across subjects $v^\text{train}=1,...,|T|$ of the training set $T$ for each threshold $t_k$ as $\text{F2}_k^\text{train} = \frac{1}{|T|} \sum_{\nu \in v^\text{train}} \text{F2}_{\nu,k}$. 
From this, we determine the optimal threshold $t_k^\text{opt}$ with $k^\text{opt}=\text{arg\,max}_k \text{F2}_k^\text{train}$ which maximizes the average $F2$ score on the training set.
We then use the optimal threshold $t_k^\text{opt}$ to obtain the mean precision, recall, and $F2$ scores across subjects. These (precision, recall, F2) on the validation or test set are our final performance metrics.

\section{Overcoming evaluation pitfalls with ALPEC}\label{appendix:evaluation_pitfalls}
Authors employing window-based evaluation often overlook reporting the class balance between arousal and non-arousal samples. In instances where the balance is disclosed, such as in the work of \cite{kuo2023machine}, who reported a ratio of 42,311:33,479 (arousals vs non-arousals), and \cite{badiei2023novel}, whose confusion matrices implied a ratio of about 1:2, discrepancies arise. Our Comprehensive Polysomnography (CPS) dataset indicates an expected ratio of about 1:5 for 30-second windows, based on the average total sleep time and the number of arousals across subjects. Such disparities are problematic for comparative analyses and from a production standpoint, as they likely lead to underestimations of false positives when background samples are underrepresented. Our ALPEC framework addresses this by sampling at the subject level rather than the window level, ensuring that validation samples are representative of the overall dataset.

Moreover, the lack of cross-subject validation is a frequent oversight with window-based evaluations, where samples from all subjects are often mixed across training, validation, and test sets. Since production models are applied to unseen subjects, it is critical to evaluate these models on new subjects during development. This practice is not consistently reported, which can inflate perceived model performance. ALPEC inherently avoids this issue by enforcing subject-level sampling, ensuring that the division of training, validation, and test samples maintains subject integrity. This approach enhances the comparability of results across studies and provides a more authentic evaluation of model efficacy.

\section{CPS dataset details}\label{appendix:cps_dataset}
\paragraph{Inclusion and exclusion criteria}
Patients included in the dataset were aged 18 or older and referred for polysomnographic examination at a sleep laboratory. Patients undergoing diagnostic treatments in the form of positive airway pressure therapies were excluded.

\paragraph{Data extraction and preprocessing}

The data extraction involved multiple steps using the SOMNOscreen device from SOMNOmedics GmbH, capturing a broad range of physiological signals. The data was further processed using the DOMINO software from the same manufacturer, which calculated additional data channels and provided initial annotations for sleep stages and arousals, which were manually reviewed and adjusted by medical experts from NRI Medizintechnik GmbH, Germany, according to guidelines from the American Academy of Sleep Medicine (AASM)~\citep{berry2012rules}. The raw data channels were upsampled to a uniform sampling rate of 256 Hz.

All input features used in this work are described in Table~\ref{tab:psg_raw_channels} (raw measurement data), Table~\ref{tab:psg_derived_channels} (derived channels), and Table \ref{tab:annotated_events} (nominal event data).

\begin{table}[!htbp]
\setlength{\tabcolsep}{4pt}
\caption{Raw data channels}\label{tab:psg_raw_channels}
\begin{center}
\begin{tabular}{ll}
\toprule
\textbf{Channels} & \textbf{Description} \\
\cmidrule(lr){1-2}
\makecell[tl]{C4:A1, C3:A2, F4:A1, O2:A1,\\A1, A2, C3, C4, F4, O2}
& \makecell[tl]{Electroencephalogram. Single electrodes mean that this\\electrode is derived against all other electrodes.} \\
\addlinespace[3pt]
Battery & Battery voltage level \\
\addlinespace[3pt]
Motion & Movement sensor measuring patient's physical activity or motion \\
\addlinespace[3pt]
Pressure Flow & \makecell[tl]{Airflow pressure measured using oxygen nasal cannula at the\\nose and mouth} \\
\addlinespace[3pt]
Thermal Flow & Thermal airflow sensor measuring breathing flow rate \\
\addlinespace[3pt]
ECG 2 & Electrocardiogram measuring heart's electrical activity \\
EMG+, EMG-, EMG & \makecell[tl]{Electromyogram measuring skeletal muscle activity at the\\left side (-) and right side (+) of the chin} \\
\addlinespace[3pt]
\makecell[tl]{EOGl, EOGL:A1, EOGL:A2,\\EOGr, EOGr:A1, EOGr:A2} & \makecell[tl]{Electrooculogram measuring the left (l) and right (r)\\eye movements} \\
\addlinespace[3pt]
Light & Ambient light sensor measuring light exposure \\
\addlinespace[3pt]
PLMl, PLMr & \makecell[tl]{Periodic Limb Movement sensors measuring limb movements\\at the left leg (l) and right leg (r)} \\
\addlinespace[3pt]
Pleth & \makecell[tl]{Plethysmography measuring changes in blood volume at the\\tip of the ring finger of the non-dominant arm} \\
\addlinespace[3pt]
Pos. & Body position sensor. Used to derive the patient's posture \\
\addlinespace[3pt]
Pulse & Pulse rate of the pulse wave \\
\addlinespace[3pt]
\makecell[tl]{RIP.Abdom, RIP.Thrx,\\Sum RIPs} & \makecell[tl]{Respiratory Inductance Plethysmography sensors measuring\\abdominal and thoracic movements during breathing.\\\textit{Sum RIPs} is a combination of \textit{RIP.Abdomen} and \textit{RIP.Thrx}} \\
\addlinespace[3pt]
SPO2 & Pulse oximetry sensor measuring blood oxygen saturation levels \\
\addlinespace[3pt]
Snoring Sound & Snore sensor measuring snoring sounds or vibrations \\
\addlinespace[3pt]
Snoring Pressure & \makecell[tl]{Pressure sensor measuring snoring intensity using oxygen nasal\\cannula at the nose and mouth} \\
\bottomrule
\end{tabular}
\end{center}
\end{table}

\begin{table}[!htbp]
\setlength{\tabcolsep}{4pt}    
\caption{Derived signals which are calculated by the DOMINO Software from the raw data}\label{tab:psg_derived_channels}
\begin{center}
\begin{tabular}{ll}
\toprule
\textbf{Signal name} & \textbf{Description} \\
\cmidrule(lr){1-2}
Syst & Systolic blood pressure curve \\
Diastol & Diastolic blood pressure curve \\
MAP & Mean arterial pressure \\
Diastol PTT & Diastolic pulse transit time \\
Systol PTT & Systolic pulse transit time \\
SpO2 & Average oxygen saturation level \\
Integrated EMG & Integrated electromyography signal from the chin \\
PTT Raw & Pulse transit time \\
HRV LF & Low frequency component of heart rate variability \\
HRV HF & High frequency component of heart rate variability \\
Heart rate & Heart rate curve \\
RR Interval & RR interval for heart rate analysis \\
SVB & Sympathovagal balance of sympathetic and parasympathetic activity \\
RR & Respiratory rate per minute \\
Obstruction & Obstruction curve in synchronized effort from abdomen and thorax \\
Phase Angle & Phase angle of synchronized effort \\
Alpha+Beta FFT & Alpha and beta wave frequency analysis in sleep \\
Delta FFT & Delta wave frequency analysis in sleep \\
Sigma FFT & Sigma wave frequency analysis in sleep \\
Average Frequency Value & Average frequency value in sleep FFT analysis \\
Activity & Activity level \\
Light & Light intensity in lux \\
\bottomrule
\end{tabular}
\end{center}
\end{table}

\begin{table}[!htb]
\caption{Annotated events that are used in this work. For a complete list of all annotated events, refer to the CPS dataset documentation~\citep{kraft2024cps}}\label{tab:annotated_events}
\begin{center}
\begin{tabular}{ll}
\toprule
\textbf{Event name} & \textbf{Description} \\
\cmidrule(lr){1-2}
Respiratory Arousal (EEG) & EEG arousal due to respiratory effort \\
Respiratory Arousal & Arousal due to respiratory effort \\
Flow Limitation Arousal (EEG) & EEG arousal due to flow limitations \\
Flow Limitation Arousal & Arousal due to flow limitations \\
SpO2 Arousal (EEG) & EEG arousal due to oxygen desaturation \\
LM Arousal (EEG) & EEG arousal due to limb movements \\
LM Arousal & Arousal due to limb movements \\
PLM Arousal (EEG) & EEG arousal due to periodic limb movements \\
PLM Arousal & Arousal due to periodic limb movements \\
Snoring Arousal (EEG) & EEG arousal due to snoring \\
Snoring Arousal & Arousal due to snoring \\
Arousal (EEG) & Spontaneous EEG arousal \\
Arousal & Spontaneous arousal \\
\addlinespace[5pt]
Sleep Profile: N1 & N1 sleep stage \\
Sleep Profile: N2 & N2 sleep stage \\
Sleep Profile: N3 & N3 sleep stage \\
Sleep Profile: Rem & Rapid Eye Movement sleep stage \\
Sleep Profile: Wach & Awake state during the measurement \\
\addlinespace[5pt]
Body Position: Prone & Prone body position \\
Body Position: Upright & Upright body position \\
Body Position: Left & Lying on the left side \\
Body Position: Right & Lying on the right side \\
Body Position: Supine & Supine body position \\
Hypopnea & Hypopnea event\\
Apnea & Apnea event\\
Central Apnea & Central apnea event \\
Obstructive Apnea & Obstructive apnea event \\
\bottomrule
\end{tabular}
\end{center}
\end{table}

Additional preprocessing of the data channels before release involved shifting the day, month, and year of all recordings to January 1, 1970, to ensure patient anonymity and converting from the European Data Format (EDF) to the Waveform Database (WFDB) format.

The target arousal classes are listed in Table~\ref{tab:annotated_events}. The presence of the postfix \textit{(EEG)} at an arousal event class indicates that the arousal was first recognized in the EEG channel, followed by its causative occurrence. In contrast, the lack of \textit{(EEG)} denotes that the causative event preceded the observable EEG effects. 
Another class of arousals that is also annotated but not included in this work are autonomic arousals. This exclusion is based on the distinct nature of autonomic arousals, which involve involuntary physiological responses regulated by the autonomic nervous system, differing from arousals typically detected in sleep studies through EEG or related to specific sleep disturbances. Autonomic arousals may not directly correlate with sleep architecture changes or the specific arousal events typically analyzed in sleep medicine, thus requiring separate consideration from a sleep medical perspective. Autonomic arousals are also typically not included in other ML-based works which focus on the general arousal detection task.

\paragraph{Dataset statistics and analysis of representativeness}
Table~\ref{tab:dataset_statistics} provides an overview of the CPS dataset, including demographic information, sleep architecture, and sleep disorder indices. Further dataset statistics are available via the supplementary material (Appendix~\ref{appendix:supp}).

\newcommand{\indentcell}[1]{\hspace{5mm}#1}

\begin{table}[!htb]
\caption{Characteristics of the CPS dataset. The total number of patients is 113. The number of patients who indicated their gender in the questionnaire was 62, no other genders were mentioned. The number of patients with REM sleep was 110.\\
Data are expressed as absolute values or as mean values $\pm$ standard deviation in the units provided.\\
OSA Severity is determined based on the AHI~\citep{wetter2012elsevier}, where \textit{Very Severe} is an additional category reserved for patients with \textit{Severe} OSA and additionally a high hypoxemia burden and high daytime sleepiness with a tendency to fall asleep during the day. The Baveno Classification for OSA severity~\citep{randerath2018challenges} was newly introduced during the collection of the CPS dataset.\\
Abbreviations: BMI: Body Mass Index, TST: Total sleep time; WASO: Wake time after sleep onset; Sleep Stages: N1, N2, N3: Stages of non-rapid eye movement sleep, REM: Rapid eye movement sleep; OSA: Obstructive sleep apnea; AHI: Apnea–hypopnea index; ArI: Arousal index; ESS: Epworth Sleepiness Scale; SPO\textsubscript{2}: Oxygen saturation level; ODI: Oxygen desaturation index ($\geq$ 3\%); T90: Time percentage spent below 90\% oxygen saturation during sleep.}\label{tab:dataset_statistics}    
\begin{center}
    \begin{tabular}{lr|lr}
    \toprule
        \textbf{Age (years)} & ~ & \textbf{OSA Severity} & ~ \\
        \indentcell{\textless 50} & 26 (23.01\%) & \indentcell{Mild} & 10 \hphantom{0}(8.85\%) \\ 
        \indentcell{50-60} & 34 (30.09\%) & \indentcell{Moderate} & 39 (34.51\%) \\ 
        \indentcell{60-70} & 23 (20.35\%) & \indentcell{Severe} & 22 (19.47\%) \\ 
        \indentcell{\textgreater70} & 22 (19.45\%) & \indentcell{Very Severe} & 17 (15.04\%) \\ 
        \indentcell{Unknown} & 8 \hphantom{0}(7.08\%) & \indentcell{Other} & 25 (22.12\%) \\ \cmidrule{1-4}
        \textbf{BMI (kg/m\textsuperscript{2})}& ~ & \textbf{Baveno Classification} & ~ \\
        \indentcell{18.5-25} & 19 (16.81\%) & \indentcell{Type A} & 15 (13.27\%) \\ 
        \indentcell{25-30} & 43 (38.05\%) & \indentcell{Type B} & 32 (28.32\%) \\ 
        \indentcell{\textgreater30} & 48 (42.48\%) & \indentcell{Type C} & 8 \hphantom{0}(7.08\%) \\ 
        \indentcell{Unknown} & 3 \hphantom{0}(2.65\%) & \indentcell{Type D} & 1 \hphantom{0}(0.88\%) \\ \cmidrule{1-2}
        \textbf{Gender} & ~ & \indentcell{Unknown} & 57 (50.44\%) \\ \cmidrule{3-4}
        
        \indentcell{Male} & 45 (72.58\%) & \textbf{Mean ESS} & 7.83 $\pm$ \hphantom{00}4.96 \\ \cmidrule{3-4}
        \indentcell{Female} & 17 (27.42\%) & \textbf{Mean Number of Arousals} & 167.56 $\pm$ \hphantom{0}87.17 \\ \cmidrule{1-4}
        \textbf{ Sleep Architecture} & ~ & \textbf{Sleep Disorder Index} & ~ \\ 
        \indentcell{Sleep Efficiency (\%)} & 70.04 $\pm$ 15.87 & \indentcell{AHI (events/hour)} & 107.86 $\pm$ \hphantom{0}39.60 \\ 
        
        \indentcell{TST (min)}  & 435.48 $\pm$ 36.83 & \indentcell{ArI (events/hour)} & 20.95 $\pm$ \hphantom{0}10.42 \\ 
        \indentcell{WASO (min)} & 130.45 $\pm$ 69.35 &  \indentcell{Snoring Index (events/hour)} & 65.25 $\pm$ 106.53 \\ \cmidrule{3-4}
        \indentcell{N1 (\% of TST)} & 15.74 $\pm$ 11.70 & \textbf{Oximetry Parameters} & ~ \\ 
        \indentcell{N2 (\% of TST)} & 51.22 $\pm$ 11.58 & \indentcell{SPO\textsubscript{2}} (\%) & 93.75 $\pm$ \hphantom{00}1.67 \\ 
        \indentcell{N3 (\% of TST)} & 19.31 $\pm$ 11.62 & \indentcell{ODI} (events / hour) & 22.23 $\pm$ \hphantom{0}15.63 \\ 
        \indentcell{REM (\% of TST)} & 13.74 $\pm$ \hphantom{0}6.40 & \indentcell{T90} (\%) & \hphantom{00}8.04 $\pm$ \hphantom{0}11.66 \\
        \bottomrule
    \end{tabular}
    \end{center}
\end{table}

Since our dataset was collected over the course of one year during routine clinical practice, it is expected to be representative of patients undergoing polysomnographic examinations in a sleep laboratory. Our analysis reveals that close to 80\% of the patients in our study suffer from obstructive sleep apnea (OSA) of varying severity. The dataset features a male-to-female ratio of 45:17 (noting that gender information is not available for all patients), with about 70\% of the patients being over 50 years old, and over 40\% classified as obese (BMI > 30).
These characteristics align with findings in existing literature, which indicate that OSA is more prevalent among males, older individuals, and those with obesity. Such demographic patterns are well-documented in research, supporting the representative nature of our dataset~\citep{bonsignore2019sex, jehan2017obstructive, fietze2019prevalence}.

\section{Data folds}
\label{appendix:data_folds}
The split between training and validation samples is performed randomly. The test set, however, was hand-selected by an expert in arousal diagnostics whose task was to find a set of samples that is representative of patients undergoing polysomnographic examinations in a sleep laboratory. A mapping of sample IDs to folds and further details can be found in Figure~\ref{fig:data_folds}.

\begin{table}[!htb]
\caption{Lists of subject IDs over data folds for the CPS dataset as used in our experiments and suggested for future work on the CPS dataset. The following entries are excluded in our experiments: Five entries marked with \dag in the training and validation folds, since they do not contain \textit{Diastol} and \textit{Syst} derived channels and one entry marked with \dag in the test set, since it contains overlapping target annotations.}
\label{fig:data_folds}    
\centering

\begin{adjustbox}{valign=t,minipage=.33\linewidth}
\captionof*{table}{(a) Training set part 1}
\vspace{1em}
\scriptsize
\begin{tabular}{l}
\texttt{3DquDEk2YwjfckxNBAQuVTshrK3VWqO7} \\
\texttt{HvVu33fnVKDLLjwY8Mtytcgi8Btsr1kS} \\
\texttt{INzmELsQB5yeF6HnHRM76U1ufVy7vmfb} \\
\texttt{hOwipKAoUqK6vJDqsjnchMKZf0e9uSH8} \\
\texttt{5Nl24yHOnojhw7nsgC0e530b2RBzOuLA} \\
\texttt{1RLVk0ocGDZLI8RRhPglAc4I3gMSLqvu} \\
\texttt{0Ah95Qw18puf1JsnrKBA6u8XXZLlMIQJ}\dag \\
\texttt{Bmy6KwUhfqRDp6bzRx1PaWoQvBpImF01} \\
\texttt{FddiLTFWMZFHH5s1NFddllezef4BJhwS} \\
\texttt{tfAnzkFia5hzaA6bHYFpkj3jPF90FzAj} \\
\texttt{tU3dZpxIdmbr9wpPpFeZGh5MciOB1TgT} \\
\texttt{FPSnBoS217CEJ8cZS7MO3VuYUJwIt8LV} \\
\texttt{5wPINWASVdhb63RK4DtJt5LuyuaWyMo8} \\
\texttt{JyXyuQIuFyAtKL8ZoWS98xvpW4PJFco6} \\
\texttt{KIdPVCRCkXIaWDQ4c4gU38xpH5PAnOSV} \\
\texttt{CMTsL0EWEJvQqMe0GKKCKN87IXp9L0UU} \\
\texttt{0pdaTB9613iRUlhUJRAYvKmMQcOV3TYn} \\
\texttt{1gQ3otWoJ3qNNJ5g4N1WtTC4JTFOlP9B} \\
\texttt{wl5eUqFCFGibvn13l8axb82mNOkp0doc} \\
\texttt{DlUvX2vg1c4ORh63BRFaCxbMlr8DeCbO} \\
\texttt{80LORVUBBBcTBzaPxmmnfrlXC3bu3Dzw} \\
\texttt{rxODabL5H6LkFhXhov0iCYKaxTA9SKSY} \\
\texttt{MYzsHdeN4rEc6ne1SobyUoK2u2bedJUp} \\
\texttt{sUDb7jmMMO6h7QCGGbkwa7LRv1JYu8hy} \\
\texttt{5C1M33g2KLtBshvnj3V6S2MYFxbvFgbr} \\
\texttt{S8dOGQgMx9WOH90orySWSfGuBL2mpxgC} \\
\texttt{UldfKBDGVlNpqUbcVPLV68eOFqucUj06} \\
\texttt{zjSqovtw62tjDA2tSelpFpAvfXbGUyI5} \\
\texttt{dxAGfKahLvCc0GhMyac32CmQo10JGLuF} \\
\texttt{3rt8nhT9Ddda15KAPVhRXJgPcilmGipU} \\
\texttt{5mhIirR785Vve6LjBZyOqRviHmWUZl9M} \\
\texttt{kxrBCJhec2Aub2FmnrU1dCIx7f3HlST6} \\
\texttt{rYbtzQzJJVg8Wksx9dU8wHg2X1R8zulJ} \\
\texttt{yAP2PJs1dDSFdYXe6GrQtQC7iO8oyX3L} \\
\end{tabular}
\end{adjustbox}%
\begin{adjustbox}{valign=t,minipage=.33\linewidth}
\captionof*{table}{(b) Training set part 2}
\vspace{1em}
\centering
\scriptsize
\begin{tabular}{l}
\texttt{hrBQwXe1RNG3VJXIAQAgC7JAjl2XlHyD} \\
\texttt{PK0t8NcGbxRcfL8tTxbrdxJe6zY1WS9f} \\
\texttt{FRU8DMa3f1esZLFfzixxhgJk0KvG7vXX} \\
\texttt{8XdpUGiGQTqm1q6E1BiNZmYTcFwQovHP} \\
\texttt{1vxo6QPCl4cn7JrPoKgTFUd6zZ6jrN6X} \\
\texttt{HeeJJy8N63XTT1mmruOcaXwl9gH06LLR} \\
\texttt{MyE1OSaDFTs03JE5K1CLusYomSaDFufz} \\
\texttt{l1VJ9A1olk859kfiGM7UNdjqitFHl3fa} \\
\texttt{eX0CkhjYbsXlnwQwAHsFDo9XyPV9b7oj} \\
\texttt{ZizO4wnchcm4tTOACaPUrkRtopGIjkFq} \\
\texttt{P45k7fhnHTEWYkEMfEPSAK2tp2hizA7O}\dag \\
\texttt{mkcvU3fDfRGULYMwqm0FCUlqPTzrSb8P} \\
\texttt{RhDBHQQCPFEvVYagp1SorSLbEygFUiAL} \\
\texttt{C2Lt7JOOpGRwdu2TrCMDNn1jE8nBCHpb} \\
\texttt{KkgHbcRejP3vgmwPVpI3jW3Pq6cddRsJ} \\
\texttt{F8Uu1bz0NcullMqx3o7S1o3M4VRg7EtR} \\
\texttt{52JOaQM1ksTl9M7xo3AYi3Hto6exeWpG} \\
\texttt{nozzxBTVeanjyN8aDAOknM5gv55sydOG} \\
\texttt{t1dRq19eE7PizNhNBod8AT5pX18KFAul} \\
\texttt{Yp6NRNtFSjbjdcFWN7Q1ATwOir9wrBNw} \\
\texttt{Oa7cBsB7PQPy7Kr8jHkJorml61mv4kIP} \\
\texttt{srARDl4a3Z4Gtb39GTOv0ynNVE6T7xHS} \\
\texttt{KX83zwqAkUKsTWQKeWmfjewObf8y7upu} \\
\texttt{wuxfEVb6iJ7GghVUjN5eKTO0VsulXaOg} \\
\texttt{gm7PhGPwaWaEeOoG0lK0altS9tLOBYGt} \\
\texttt{zWtiCjFSxBFRmU3DklC4UMFKFHCOXJgS} \\
\texttt{xc54eJI7x5BwwtOLSfdRnYdvMLagr1sp} \\
\texttt{RqC1HNwuWs5fhqEheSYlRc93EoReggVm} \\
\texttt{U2onxpoiCTT2F4SHmDMTeWb2GgWtRZhb} \\
\texttt{Adm1jCIZpVR5Evov0BZ5XyEA3QXgkpiO} \\
\texttt{niVBznqEcE3RfyWa2u7EXjpguXBMB9dE} \\
\texttt{nrtZyE7IBmlm0ZH3gG7FHJAoYxlFqczB}\dag \\
\texttt{x6nbOt0fqHhNHbYCgQP0Eaa7ymp6eKuq} \\
\texttt{QHauJpImWK54mPysmthbCSRI6BwQfuim} \\
\end{tabular}
\end{adjustbox}%
\begin{adjustbox}{valign=t,minipage=.33\linewidth}
\captionof*{table}{(c) Validation set}
\vspace{1em}
\scriptsize
\begin{tabular}{l}
\texttt{I4PBCtY88EMiljTpJ6ns5kIoimlIUgfl} \\
\texttt{ip25KNFw4RbDNeQTXjLI95OsnQLLe35e} \\
\texttt{vvcikOyQkXvfZdNkFYEWveZjUlQmYrSf} \\
\texttt{euzXQcpDnB1xsKeKOJHYW48lSNZXiVHK} \\
\texttt{AavMakHhFeoHz9AgdWVTBsXCdiOLBNEk} \\
\texttt{KBn6Fz4XRNh5A7YRBBGZIlSVzcmV6P51} \\
\texttt{AEhNZrB5mb9K7hCBxB0xvJzUC5WxqaFN} \\
\texttt{ssRHUDqjbEgtepnEnjPpOd6WQHekD6PZ} \\
\texttt{XpSmjNEW0Mad8655KQ4q3NjDHNUNHW6C} \\
\texttt{9q7wGfr4xVuiodCCrqViuS1dnZ8tthvZ} \\
\texttt{A2YTNgCrkIoMGvSkyzK5R0EFgXVRfptP} \\
\texttt{xijlZSyZXb5MEfbKLM54iOJRJkRdrBrX} \\
\texttt{OCi1DrQQgJ7GjNFNTrxscclxkaDH0Y76} \\
\texttt{yWLp2YwYQlKqoSR14JUiazZzcywX8Xj8} \\
\texttt{jusHjPORbNBFg6iMvaNE4HcFMc4oJyqD} \\
\texttt{3veUj2KxK6jmJlllRj9tXbSncK6xnS0x} \\
\texttt{ZwojnSyDEhD6s8ZERY8AoDWL0cnBH7BU} \\
\texttt{GJShIRZ3lsuqeG7HhgpwlMw5v8prIuD4} \\
\texttt{ZFkXoBHziu00Toye78g1YdKm5P0bA0dY} \\
\texttt{pY4EUZQszT0kL5Et76FkEbRlLeAMT2hz} \\
\texttt{oTlKy2ISbZfN7i2jTcZ4mCqmCpK6dhDm} \\
\texttt{TLYaLIWCrBbtqLySvqxj67g5ZAfn2Zhs}\dag \\
\texttt{MT9Dkn3L0akMVov0lRUq9HmNsP49R1dX} \\
\texttt{rHg2gQNoevGYPUag6PAn9CANXKmx17ms} \\
\texttt{Vth5VKxswvoQEpLCis200xjS0f0ctjFu} \\
\texttt{vc3ShhPeF5CiTm9Hi0moakicyNWNulab} \\
\texttt{XSedsGunPd3IUuZ8RJUGmz3SUoCv0rzW} \\
\texttt{Pk9lFIEExot3gjvV83ZDTuWSj6dqk9uW} \\
\texttt{tihrah4T4i2gA8dscroj9Mu5715fXojg} \\
\texttt{M7d3C4ZQtx0R00Wum7JMxQtZExUfNEzi}\dag \\
\end{tabular}
\end{adjustbox}%
\vspace{0.5em}
\begin{adjustbox}{valign=b,minipage=.49\linewidth}
\captionof*{table}{(d) Test set}
\vspace{1em}
\centering
\scriptsize
\begin{tabular}{l}
\texttt{vHJMSYFIl1TfLweQ5DWMGN5f47ULFNxe} \\
\texttt{MT1zW5iB0h1bxF42QBpyDqotQk7NcnHw} \\
\texttt{leySrSnra9yAO3eTJIGB55nrjRS3RqIW} \\
\texttt{FLsgQZoIGHx1G3LmdD7jtICMik2EKRKN} \\
\texttt{YFQX33c8EEoapTndd2084KbUuUmtj7xF} \\
\texttt{KB84bUmLWOrKCKkISCn8QuNBhF5mg0L8} \\
\texttt{oEJ7fslCTL7s0OfIe7nYIPqo7Il4rMjI} \\
\texttt{CzwqE37s81YahjNICSXI2Tb4Fmp6bclE} \\
\texttt{Su02hndUSYGSKJmcSqroKmtDjXIJ4y60} \\
\texttt{LcsapTberZwzU7qyEr11andO59HTOVCv} \\
\texttt{0vjSLbBj2sckqQam3tZ92QLDpQNYqaa0}\dag \\
\texttt{mXHZZ887A9fcZgOmnxhnPVHwu5ECljDG} \\
\texttt{RpARZ17l5osnFUcqIj2aTOsgRBMDutoA} \\
\texttt{tIgyhF8T1BOZnu7h6jb58igU5MAGgdo9} \\
\texttt{kKDzUlAprXqDz84Nrw9UP1W0jpgUKkhN} \\
\end{tabular}
\end{adjustbox}%
\end{table}

\section{Supplementary Material}\label{appendix:supp}

\subsection{Access to the dataset}\label{sec:access}

The CPS dataset~\citep{kraft2024cps} is accessible on the PhysioNet platform at \url{https://doi.org/10.13026/sxs0-h317} under the PhysioNet Credentialed Health Data License 1.5.0.

\subsection{Datasheet}\label{sec:datasheet}

This datasheet is based on the \textit{Datasheets for Datasets} framework~\cite{gebru2021datasheets}.

\subsubsection{Motivation}

\question{For what purpose was the dataset created? Was there a specific task in mind? Was there a specific gap that needed to be filled? Please provide a description.}\\
The CPS dataset was compiled to conduct a clinical study on arousal diagnostics (see \url{https://drks.de/search/en/trial/DRKS00033641​}). The primary study goals are to investigate if Machine Learning can enhance the quality and efficiency of sleep-related arousal diagnostics, while also reducing technical demands. The dataset was created with the specific task of refining the diagnostic workflow by leveraging Photoplethysmography (PPG) data to reduce reliance on comprehensive EEG, electromyography (EMG), and electrooculography (EOG) inputs.

\question{Who created the dataset (e.g., team, research group) and on behalf of which entity (e.g., company, institution, organization)?}\\
The dataset was compiled from patients undergoing regular polysomnographic examinations at the sleep laboratory of the Klinik für Kardiologie, Pneumologie und Angiologie at Klinikum Esslingen. The companies IT-Designers Gruppe and NRI Medizintechnik GmbH were involved in collecting and processing the dataset. IT-Designers Gruppe initiated, funded, and supported the research. NRI Medizintechnik GmbH operates the sleep laboratory and cooperated and assisted in implementing the data collection protocol. Technical support was provided by SOMNOmedics GmbH, the supplier for the hardware and software of the sleep laboratory. The clinic and companies are all based in Germany.

\question{Who funded the creation of the dataset? If there is an associated grant, please provide the name of the grant and the grant ID.}\\
This research was funded by STZ Softwaretechnik GmbH, part of the IT-Designers Gruppe, Esslingen am Neckar, Germany. It contains all samples that have been collected during the clinical study.

\question{Any other comments?}\\
None.

\subsubsection{Composition}

\question{What do the instances that comprise the dataset represent (e.g., documents, photos, people, countries)? Are there multiple types of instances (e.g., nodes, edges) present in the dataset? Please provide a description.}\\
The instances in the dataset represent diagnostic polysomnographic sleep recordings, which include up to 36 raw and 23 derived data channels, alongside 81 types of annotated events for each participant, supplemented by data from various questionnaires.

\question{How many instances are there in total (of each type, if appropriate)?}\\
The dataset encompasses 113 diagnostic polysomnographic sleep recordings.

\question{Does the dataset contain all possible instances or is it a sample (not necessarily random) of instances from a larger set? If the dataset is a sample, then what is the larger set? Is the sample representative of the larger set (e.g., geographic coverage)? If so, please describe how this representativeness was validated/verified. If it is not representative of the larger set, please describe why not.}\\
The dataset is a sample of diagnostic sleep recordings from adult patients undergoing regular and purely diagnostic examinations at the sleep laboratory of the Klinik für Kardiologie, Angiologie und Pneumologie of the medical clinic in Esslingen am Neckar, Germany. It contains all samples that have been collected during the clinical study. The representativeness compared to other datasets on arousal diagnostics is discussed in the appendix of the main part of this publication.

\question{What data does each instance consist of? “Raw” data (e.g., unprocessed text or images) or features? In either case, please provide a description.}\\
Each instance consists of raw polysomnographic data including up to 36 raw and 23 derived data channels, 81 types of annotated events, and data from various questionnaires. Descriptions of all channels and fields that are used in this work are provided in the appendix of the main part of this publication. A full description of the dataset is available on the PhysioNet page of the CPS dataset~\citep{kraft2024cps}.

\question{Is there a label or target associated with each instance? If so, please provide a description.}\\
Yes, each recording includes labels for sleep-related events such as arousals, apnea, hypopnea, and other sleep events as per the American Association of Sleep Medicine (AASM) guidelines~\citep{berry2012rules}. A full list of annotated events is available on the PhysioNet page of the CPS dataset~\citep{kraft2024cps}.

\question{Is any information missing from individual instances? If so, please provide a description, explaining why this information is missing (e.g., because it was unavailable). This does not include intentionally removed information, but might include, e.g., redacted text.}\\
The Pittsburgh Sleep Quality Index (PSQI) questionnaire was only given to 62 patients. The remaining 51 patients did not receive the questionnaire, so this information is missing for those instances. All questionnaires contain missing values for some questions due to non-response. Raw data is complete for all patients, but five patients are missing derived systolic and diastolic blood pressure channels.

\question{Are relationships between individual instances made explicit (e.g., users’ movie ratings, social network links)? If so, please describe how these relationships are made explicit.}\\
No explicit relationships between individual instances are made.

\question{Are there recommended data splits (e.g., training, development/validation, testing)? If so, please provide a description of these splits, explaining the rationale behind them.}\\
We provide recommended splits for training, validation, and test data, listed in the appendix of the main part of this publication and on the PhysioNet page of the CPS dataset~\citep{kraft2024cps}.

\question{Are there any errors, sources of noise, or redundancies in the dataset? If so, please provide a description.}\\
The dataset may contain some noise typical of polysomnographic data, such as artifacts from patient movement or external interference, but efforts were made to minimize these by conducting comprehensive quality assurance in the pilot phase of the clinical study. The channels \textit{Heart Rate}, \textit{Light}, and \textit{SpO2} are smoothed versions of the raw data channels \textit{Pulse}, \textit{Light}, and \textit{SPO2}, respectively, processed using the DOMINO software from SOMNOmedics GmbH, Germany.

\question{Is the dataset self-contained, or does it link to or otherwise rely on external resources (e.g., websites, tweets, other datasets)? If external resources are required, please describe them, as well as any restrictions (e.g., licenses, fees) associated with them.}\\
The dataset is self-contained.

\question{Does the dataset contain data that might be considered confidential (e.g., data that is protected by legal privilege or by doctor-patient confidentiality, data that includes the content of individuals’ non-public communications)? If so, please provide a description.}\\
Yes, the dataset contains confidential patient data protected under doctor-patient confidentiality agreements. The most sensitive data are the sleep medical diagnoses. The dataset has been anonymized to protect patient identities.

\question{Does the dataset contain data that, if viewed directly, might be offensive, insulting, threatening, or might otherwise cause anxiety? If so, please describe why.}\\
No, the dataset does not contain such data.

\question{Does the dataset identify any subpopulations (e.g., by age, gender)? If so, please describe how these subpopulations are identified and provide a description of their respective distributions within the dataset.}\\
The dataset includes demographic information such as age and gender to study their impact on sleep-related arousals. It was determined from questionnaires. The age distribution in years is: <50: 26 patients (23.01\%), 50-60: 34 patients (30.09\%), 60-70: 23 patients (20.35\%), >70: 22 patients (19.45\%), unknown: 8 patients (7.08\%). The approximate gender distribution, based on 62 patients, is: Male: 45 patients (72.58\%), Female: 17 patients (27.42\%). Gender information is only available in the aggregated statistics, not for individual patients, as an anonymization measure.

\question{Is it possible to identify individuals (i.e., one or more natural persons), either directly or indirectly (i.e., in combination with other data) from the dataset? If so, please describe how.}\\
No, patient identities are anonymized to prevent identification.

\question{Does the dataset contain data that might be considered sensitive in any way (e.g., data that reveals racial or ethnic origins, sexual orientations, religious beliefs, political opinions or union memberships, locations of health data about individuals or genetic data, forms of financial information, such as social security numbers, salary)? If so, please provide a description.}\\
The most sensitive data are the sleep medical diagnoses, consisting of short extracted textual descriptions of patients' sleep-related medical conditions from doctors' letters, e.g., obstructive sleep apnea with severity indications. Statistics and a complete listing of the sleep medical diagnoses are available on the PhysioNet page of the CPS dataset~\citep{kraft2024cps} and in the file \textit{statistics.html} in this supplementary material (generated using the SweetViz~\citep{bertrand2020sweetviz} library, under MIT license).

\question{Any other comments?}\\
None.

\subsubsection{Collection Process}

\question{How was the data associated with each instance acquired? Was the data directly observable (e.g., raw text, movie rating), reported by subjects (e.g., survey responses), or indirectly inferred/derived from other data (e.g., part of speech tags, model-based guesses for age or language)? If the data was reported by subjects or indirectly inferred/derived from other data, was the data validated/verified? If so, please describe how.}\\
The data was directly observable from polysomnographic recordings and supplemented by data from various questionnaires.

\question{What mechanisms or procedures were used to collect the data (e.g., hardware apparatuses or sensors, manual human curation, software programs, software APIs)? How were these mechanisms or procedures validated?}\\
The data collection during polysomnographic examinations involved SOMNOscreen devices and the DOMINO software, both from SOMNOmedics GmbH, Germany. The data was validated, curated, and extended with additional labels (e.g., sleep stages and arousals) by trained sleep medical scorers from NRI Medizintechnik GmbH (Germany) following guidelines from the American Academy of Sleep Medicine (AASM). The data was exported from DOMINO in EDF (raw data) and TXT (annotations) formats. An employee of Klinikum Esslingen (funded by IT-Designers Gruppe) performed the digitalization of the questionnaires and doctor's letter in YAML-format, the pseudonymization of the whole data and made the data available to the research group from IT-Designers Gruppe. The data was then further anonymized for release. The whole process was developed in collaboration with data protection officers from Klinikum Esslingen and IT-Designers Gruppe. It was approved by the ethics committee of the Landesärztekammer Baden-Württemberg, Germany. The data quality was validated in a pilot phase.

\question{If the dataset is a sample from a larger set, what was the sampling strategy (e.g., deterministic, probabilistic with specific sampling probabilities)?}\\
The dataset was collected from a monocentric study at the sleep laboratory of the Klinik für Kardiologie, Angiologie und Pneumologie in Esslingen am Neckar, Germany, ensuring a representative sample of adult patients undergoing regular diagnostic examinations.

\question{Who was involved in the data collection process (e.g., students, crowdworkers, contractors) and how were they compensated (e.g., how much were crowdworkers paid)?}\\
The data collection was conducted by clinical staff at Klinikum Esslingen. One student was hired by Klinikum Esslingen to perform the digitalization and pseudonymization of the data. He was paid 12.98 EUR per hour. The expenses for the student were covered by IT-Designers Gruppe.

\question{Over what timeframe was the data collected? Does this timeframe match the creation timeframe of the data associated with the instances? If not, please describe the timeframe in which the data associated with the instances was created.}\\
The data was collected during 2021-2022.

\question{Were any ethical review processes conducted (e.g., by an institutional review board)? If so, please provide a description of these review processes, including the outcomes, as well as a link or other access point to any supporting documentation.}\\
The study protocol was approved by the ethics committee of the Landesärztekammer Baden-Württemberg on 2020-10-21 (committee number F-2020-105, \url{https://www.aerztekammer-bw.de/ethikkommission}). The clinical study was registered at the German Clinical Trials Register, DRKS-ID: DRKS00033641 (\url{https://drks.de/search/en/trial/DRKS00033641​}).

\question{Does the dataset relate to people? If not, you may skip the remaining questions in this section.}\\
Yes, the dataset relates to people.

\question{Did you collect the data from individuals in question directly, or obtain it via third parties or other sources (e.g., websites)?}\\
The data was collected directly from individuals at the Klinikum Esslingen sleep laboratory.

\question{Were the individuals in question notified about the data collection? If so, please describe (or show with screenshots or other information) how notice was provided, and provide a link or other access point to, or otherwise reproduce, any supporting documentation.}\\
Patients were informed as part of the clinical study consent process. The consent form followed a template and was approved by the ethics committee of the federal state, the Landesärztekammer Baden-Württemberg, Germany.

\question{Did the individuals in question consent to the collection and use of their data? If so, please describe (or show with screenshots or other information) how consent was requested and provided, and provide a link or other access point to, or otherwise reproduce, any supporting documentation.}\\
Yes, patients gave informed consent for the collection and use of their data. The content of the clinical study and all general conditions were explained verbally by the medical staff and in writing in the informed consent form. It was administered in a preliminary visit prior to the examination.

\question{If consent was obtained, were the consenting individuals provided with a mechanism to revoke their consent in the future or for certain uses? If so, please provide a description, as well as a link or other access point to, or otherwise reproduce, any supporting documentation.}\\
Patients were informed of their right to revoke consent at any time.

\question{Has an analysis of the potential impact of the dataset and its use on data subjects (e.g., a data protection impact analysis) been conducted? If so, please provide a description of this analysis, including the outcomes, as well as a link or other access point to any supporting documentation.}\\
Yes, an impact analysis was conducted with a focus on the risks of re-identification, data misuse, data breaches, and failure to achieve the study objectives. Outcomes (measures) to reduce the risks were as follows: Selection of an established platform for data sharing (PhysioNet) including usage limitations (future usage must be in line with the original study goals) and the requirement of a data use agreement and public credentialed access. Additionally, in order to anonymize the data, we removed most free text, the sensitive attributes medication and pre-existing conditions, and multiple indirect identifiers like gender and profession that were less important for achieving the study goals. For the remaining indirect identifiers (age and BMI), we selected bins that respected k-anonymity with k=3. For the medical diagnosis (the remaining sensitive attribute), we made sure to have i-diversity with i=2 among the k-anonymous groups. Absolute timestamps in the measurement data were adjusted to start on January 1st, 1970. Our measures cover the criteria required by the safe harbor method from the Health Insurance Insurance Portability and Accountability Act (HIPAA) and go beyond them.

\question{Any other comments?}\\
None.

\subsubsection{Preprocessing/cleaning/labeling}

\question{Was any preprocessing/cleaning/labeling of the data done (e.g., discretization or bucketing, tokenization, part-of-speech tagging, SIFT feature extraction, removal of instances, processing of missing values)? If so, please provide a description. If not, you may skip the remaining questions in this section.}\\
Derived data channels and some event data were automatically calculated by the DOMINO software from SOMNOmedics GmbH. Manual labeling was performed by trained sleep medical scorers from NRI Medizintechnik GmbH following guidelines from the American Academy of Sleep Medicine (AASM). Bucketing of age and BMI attributes and shifting of timestamps were performed to anonymize the data. Apart from this, all raw data were converted from EDF to WFDB format, which is the standard format for PhysioNet. In the process, all raw data was upsampled to the highest sampling rate of 256 Hz.

\question{Was the “raw” data saved in addition to the preprocessed/cleaned/labeled data? If so, please provide a link or other access point to the “raw” data.}\\
Access to the raw data beyond the clinical study is limited to the medical clinic, Klinikum Esslingen.

\question{Is the software used to preprocess/clean/label the data available? If so, please provide a link or other access point.}\\
The software that was used for preprocessing is proprietary. The DOMINO software from SOMNOmedics GmbH can be licensed from the company.

\question{Any other comments?}\\
None.

\subsubsection{Uses}

\question{Has the dataset been used for any tasks already? If so, please provide a description.}\\
This publication entails the first use of the dataset.

\question{Is there a repository that links to any or all papers or systems that use the dataset? If so, please provide a link or other access point.}\\
No.

\question{What (other) tasks could the dataset be used for?}\\
The dataset may be used for research that aligns with the original study goals (see \url{https://drks.de/search/en/trial/DRKS00033641}). The study is aimed at investigating the utility of Machine Learning (ML) for improving the quality and efficiency of sleep-related arousal diagnostics, reducing technical demands of the data collection process, assessing the utility of a transparent clinical decision support system, studying the clinical relevance of arousals on sleep quality, and the utility of ML for medical knowledge discovery in this context.

\question{Is there anything about the composition of the dataset or the way it was collected and preprocessed/cleaned/labeled that might limit its usability for other tasks?}\\
We do not foresee any limitations on the usability of the dataset for the tasks mentioned above.

\question{Any other comments?}\\
None.

\subsubsection{Distribution}

\question{Will the dataset be distributed to third parties outside of the entity (e.g., company, institution, organization) on behalf of which the dataset was created? If so, please provide a description.}\\
Yes, the dataset is publicly available for research purposes with credentialed access on PhysioNet.

\question{How will the dataset be distributed (e.g., tarball on website, API, GitHub)? Does the dataset have a digital object identifier (DOI)?}\\
The dataset is available on PhysioNet at \url{https://doi.org/10.13026/sxs0-h317}.

\question{When will the dataset be distributed?}\\
The dataset is already available on PhysioNet.

\question{Will the dataset be distributed under a copyright or other intellectual property (IP) license, and/or under applicable terms of use (ToU)? If so, please describe this license and/or ToU, and provide a link or other access point to, or otherwise reproduce, any supporting documentation.}\\
The dataset is distributed under the PhysioNet Credentialed Health Data License 1.5.0 (\url{https://physionet.org/about/licenses/physionet-credentialed-health-data-license-150/}).

\question{Have any third parties imposed IP-based or other restrictions on the data associated with the instances? If so, please describe these restrictions, and provide a link or other access point to, or otherwise reproduce, any supporting documentation.}\\
There are no third-party IP-based restrictions.

\question{Do any export controls or other regulatory restrictions apply to the dataset or to individual instances? If so, please describe these restrictions, and provide a link or other access point to, or otherwise reproduce, any supporting documentation.}\\
There are no export controls or other regulatory restrictions.

\question{Any other comments?}\\
None.

\subsubsection{Maintenance}

\question{Who will be supporting/hosting/maintaining the dataset?}\\
The dataset is hosted on the PhysioNet platform. Support and maintenance will be provided by the authors of this publication.

\question{How can the owner/curator/manager of the dataset be contacted (e.g., email address)?}\\
Contact information for the dataset maintainers can be found on the PhysioNet page of the CPS dataset~\citep{kraft2024cps} under the \textit{Corresponding Author} section.

\question{Is there an erratum? If so, please provide a link or other access point.}\\
Currently, there is no erratum. If the need for an erratum arises, the dataset can be updated on PhysioNet with semantic versioning.

\question{Will the dataset be updated (e.g., to correct labeling errors, add new instances, delete instances)? If so, please describe how often, and how updates will be communicated to dataset consumers (e.g., mailing list, GitHub)?}\\
Updates will be deployed as necessary to correct any possible errors. Communication will be done via the PhysioNet website.

\question{If the dataset relates to people, are there applicable limits on the retention of the data associated with the instances (e.g., were the individuals in question told that their data would be retained for a fixed period of time, or were they told that they could have their data deleted)? If so, please describe these limits and explain how these limits will be enforced.}\\
The anonymized data will be retained indefinitely on PhysioNet. Access to the data pre-anonymization is limited to the research group from IT-Designers Gruppe for four years after the end of the data collection, which was communicated to the study participants. Klinikum Esslingen will retain the original data in accordance with legal requirements.
Requests for deletion of data only affect the pre-anonymized data. Since the patient IDs in the anonymization process were created randomly and not linked to any patient information, deletion of the anonymized data of a specific patient is technically not possible.

\question{Will older versions of the dataset continue to be supported/hosted/maintained? If so, please describe how. If not, please describe how its obsolescence will be communicated to dataset consumers.}\\
Older versions will be maintained on PhysioNet to ensure continuity and reproducibility of research.

\question{If others want to extend/augment/build on/contribute to the dataset, is there a mechanism for them to do so? If so, please provide a description.}\\
We are not aware of any mechanism on PhysioNet for users to contribute directly to the dataset. However, we are open to collaboration and will consider any requests for extensions or contributions.

\question{Any other comments?}\\
None.

\subsection{Detailed description of the dataset}\label{sec:detailed-description}
A detailed description of all channels and fields within the dataset, translations of data fields from German to English, and Croissant~\citep{akhtar2024croissant} metadata (under the Apache-2.0 license) are provided on the PhysioNet page of the CPS dataset~\citep{kraft2024cps}. To avoid redundancy and potential ambiguities in case of updates on the PhysioNet page, we have not included this information in this supplementary material. All data from the CPS dataset used in this publication is described in the main paper, primarily within the appendix.

\subsubsection{Loading the dataset}\label{sec:loading-dataset}
The documentation of the CPS dataset on PhysioNet~\citep{kraft2024cps} contains code files and instructions on how to load the dataset based on Croissant specifications~\citep{akhtar2024croissant}. These are also attached to this supplementary material. Use instructions are provided in the \textit{README.md} file.

\subsubsection{Dataset statistics}
The CPS dataset page on PhysioNet, along with this supplementary material, includes a script named \textit{generate\_statistics.py} that uses the data loading functions described in Section \ref{sec:loading-dataset} to load the data. This script generates basic demographic statistics, statistics on questionnaire answers, medical diagnoses, and additional derived statistics (\eg, distribution of the number of arousals per subject). All statistics are automatically generated using the SweetViz~\citep{bertrand2020sweetviz} library, available under the MIT license. Precomputed statistics are provided in the file \textit{statistics.html}.

\subsection{Utilized compute and environmental impact}\label{sec:compute-environment}
Experiments following the \textit{DeepSleep} approach ran on single Nvidia GeForce RTX2080TI GPUs with 11 GB of memory. They are part of a workstation containing four GPUs. The workstation is equipped with an Intel Core i9-10900X CPU with 10 cores and 20 threads and 128 GB of DDR4 RAM.
Experiments using models implemented in the \textit{sktim} library~\citep{loning2019sktime} ran on the CPU of the same workstation.

Table \ref{tab:impact} gives an overview of the average runtime for single experiments conducted for this publication, where we also indicate how often experiments were repeated to obtain confidence intervals.

\begin{table}[!htbp]
\caption{Runtime and count of experiments conducted for this publication.}
\label{tab:impact}
\begin{small}
\begin{center}
\begin{tabular}{lllll}

\toprule
& Experiment & Environment & Repetitions & Average runtime [h]\\
\cmidrule(lr){1-5}
\parbox[t]{2mm}{\multirow{14}{*}{\rotatebox[origin=c]{90}{Main experiments on CPS dataset}}}

\,\,& D4 & GPU & 5 & 10.8 \\
& D3 & GPU & 5 & 6.5 \\
& D2 & GPU & 5 & 0.7 \\
& D1 & GPU & 5 & 0.7 \\
& IndividualBOSS & CPU & 5 & 0.18 \\
& SupervisedTimeSeriesForest & CPU & 5 & 1.4 \\
& TimeSeriesForestClassifier & CPU & 5 & 0.7 \\
& SignatureClassifier & CPU & 5 & 1.1 \\
& SummaryClassifier & CPU & 5 & 0.2 \\
& Catch22Classifier & CPU & 5 & 0.4 \\
& RandomStratified & CPU & 5 & 0.03 \\
& RandomUniform & CPU & 5 & 0.03 \\
& Constant 1 & CPU & 1 & 0.03 \\
& Constant 0 & CPU & 1 & 0.03 \\

\midrule

\parbox[t]{2mm}{\multirow{6}{*}{\rotatebox[origin=c]{90}{\makecell[tl]{2018 PhysioNet\\Challenge}}}}

& RERA, IOD & GPU & 5 & 11.6 \\
& RERA, POD & GPU & 1 & 23,75 \\
& RERA, FED & GPU & 5 & 6.8 \\
& Most freq., IOD & GPU & 5 & 10,6 \\
& Most freq., POD & GPU & 1 & 33,2 \\
& Most freq., FED & GPU & 5 & 11,9 \\

\midrule

\parbox[t]{2mm}{\multirow{2}{*}{\rotatebox[origin=c]{90}{MISC}}}

& Ablation study & GPU & 30 & 0.65 \\
& HP Tuning & GPU & 44 & 0.38 \\

\bottomrule
\end{tabular}
\end{center}
\end{small}
\end{table}

In total, we have 391.17 GPU hours and 20,26 hours without GPU usage.
From this, we conduct estimations of kgCO$_2$eq for the GPU hours using the \href{https://mlco2.github.io/impact#compute}{MachineLearning Impact calculator} presented in~\citet{lacoste2019quantifying}. We use a factor of 0.4880 kgCO$_2$eq per kWh for the electricity mix in Germany. 

Total emissions are estimated to be 47.72 kgCO$_2$eq. Due to preliminary exploratory experiments and repetitions of experiments due to changes in requirements or fixes, we estimate the total emmission to be thrice as high, i.e. about 150 kgCO$_2$eq.

\subsection{Author Statement of Responsibility}\label{sec:author-statement}
The authors of this publication bear all responsibility in case of violation of rights and confirm that the dataset will be distributed under the PhysioNet Credentialed Health Data License 1.5.0.

\end{appendix}
\end{document}